%% file: main.tex
\documentclass[10pt,twocolumn,letterpaper]{article}

\usepackage[T1]{fontenc}

\usepackage{wacv}
\usepackage{times}
\usepackage{epsfig}
\usepackage{graphicx}
\usepackage{amsmath}
\usepackage{amssymb}
\usepackage{booktabs}
\usepackage{multicol}
\usepackage{siunitx}

\usepackage{adjustbox} 
\usepackage{array}
\newcolumntype{R}[2]{%
    >{\adjustbox{angle=#1,lap=\width-(#2)}\bgroup}%
    l%
    <{\egroup}%
}

\wacvfinalcopy
\usepackage[breaklinks=true,bookmarks=false,colorlinks,%
            pdftex,%
            pdfauthor={Alexander Mathis, Thomas Biasi, Steffen Schneider,
                       Mert Yüksekgönül, Byron Rogers, Matthias Bethge,
                       Mackenzie W. Mathis},%
            pdftitle={Pretraining boosts out-of-domain robustness for pose estimation},%
            pdfsubject={Pretraining boosts out-of-domain robustness for pose estimation},%
            pdfkeywords={Pose Estimation, Transfer Learning}
]{hyperref}

\newcommand*\samethanks[1][\value{footnote}]{\footnotemark[#1]}

\begin{document}

\title{Pretraining boosts out-of-domain robustness for pose estimation}

\author{%
    Alexander Mathis$^{1,2}$%
            \thanks{Equal contribution.}\;%
            \thanks{Correspondence: {\tt alexander.mathis@epfl.ch}}\quad
    Thomas Biasi$^2$\samethanks[1]\quad
    Steffen Schneider$^{1,3}$\quad
    Mert Y\"uksekg\"on\"ul$^{2,3}$\\
    Byron Rogers$^4$\quad
    Matthias Bethge$^3$\quad
    Mackenzie W. Mathis$^{1,2}$%
        \thanks{Correspondence: {\tt mackenzie.mathis@epfl.ch}}\\
{\small
    $^1$EPFL\quad
    $^2$Harvard University\quad
    $^3$University of Tübingen\quad
    $^4$Performance Genetics
}
}

\maketitle

\begin{abstract}
    Neural networks are highly effective tools for pose estimation. However, as in other computer vision tasks, robustness to out-of-domain data remains a challenge, especially for small training sets that are common for real-world applications. Here, we probe the generalization ability with three architecture classes (MobileNetV2s, ResNets, and EfficientNets) for pose estimation. We developed a dataset of 30 horses that allowed for both ``within-domain'' and ``out-of-domain'' (unseen horse) benchmarking---this is a crucial test for robustness that current human pose estimation benchmarks do not directly address. We show that better ImageNet-performing architectures perform better on both within- and out-of-domain data if they are first pretrained on ImageNet. We additionally show that better ImageNet models generalize better across animal species. Furthermore, we introduce Horse-C, a new benchmark for common corruptions for pose estimation, and confirm that pretraining increases performance in this domain shift context as well. Overall, our results demonstrate that transfer learning is beneficial for out-of-domain robustness.
\end{abstract}

\section{Introduction}

\begin{figure}[t]
    \includegraphics[width=.49\textwidth]{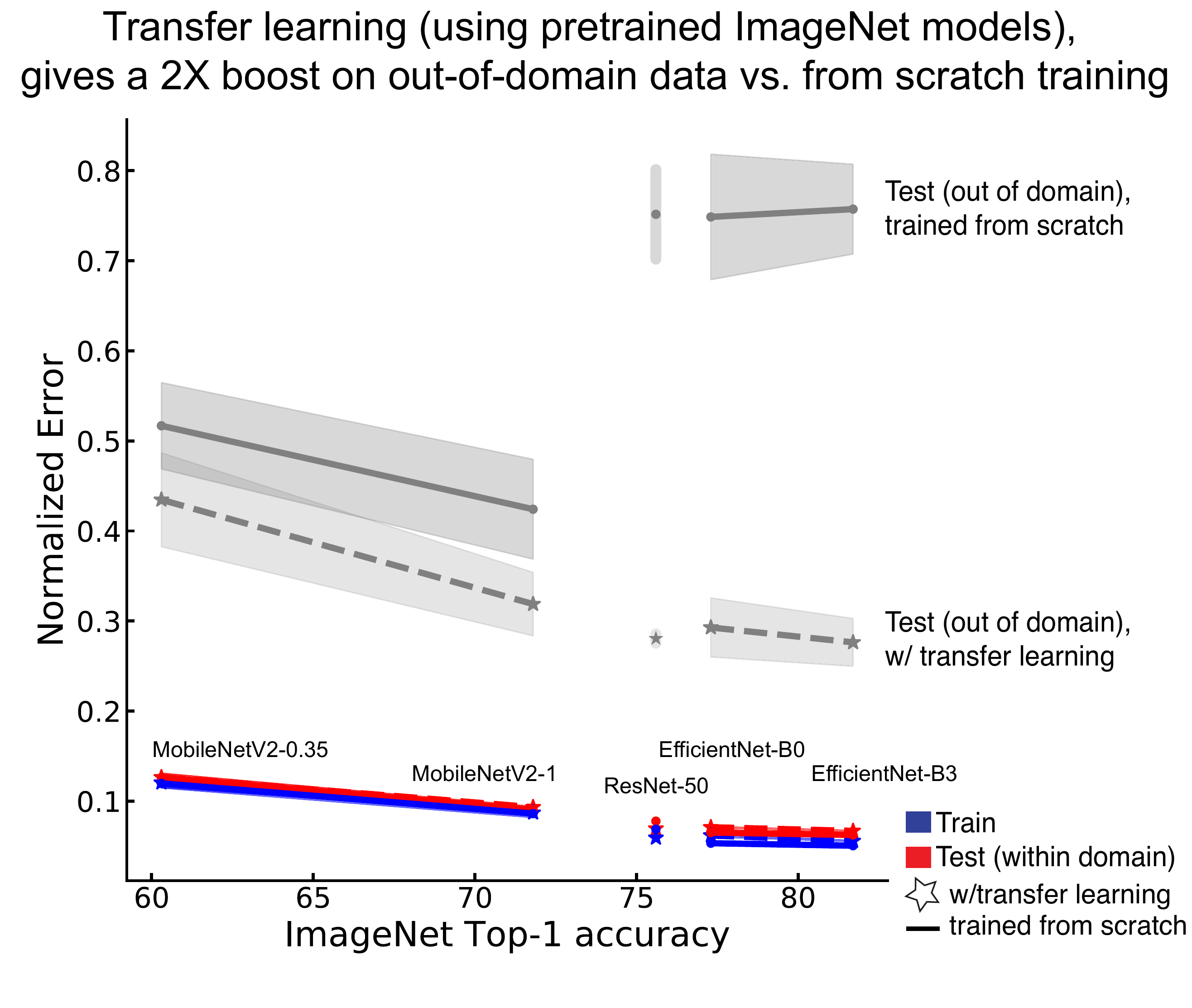}
    \caption{ {\bf Transfer Learning boosts performance, especially on out-of-domain data.} Normalized pose estimation error vs. ImageNet Top 1\% accuracy with different backbones. While training from scratch reaches the same task performance as fine-tuning, the networks remain less robust as demonstrated by poor accuracy on out-of-domain horses. Mean $\pm$ SEM, 3 shuffles.}
    \vspace{-15pt}
    \label{fig:highlight}
\end{figure}

Pose estimation is an important tool for measuring behavior, and thus widely used in technology, medicine and biology~\cite{bachmann2015evaluation, ostrek2019existing, maceira2019wearable, mathis2020deep}. Due to innovations in both deep learning algorithms~\cite{insafutdinov2017cvpr, cao2017realtime, he2017mask, kreiss2019pifpaf, ning2020lighttrack, cheng2020higherhrnet} and large-scale datasets~\cite{lin2014microsoft,andriluka20142d,andriluka2018posetrack} pose estimation on humans has become very powerful. However, typical human pose estimation benchmarks, such as MPII pose and COCO~\cite{lin2014microsoft,andriluka20142d,andriluka2018posetrack}, contain many different individuals (>10k) in different contexts, but only very few example postures per individual. In real world applications of pose estimation, users often want to create customized networks that estimate the location of user-defined bodyparts by only labeling a few hundred frames on a small subset of individuals, yet want this to generalize to new individuals~\cite{ostrek2019existing, maceira2019wearable, mathis2020deep, sanakoyeu2020transferring}. Thus, one naturally asks the following question: Assume you have trained an algorithm that performs with high accuracy on a given (individual) animal for the whole repertoire of movement---how well will it generalize to different individuals that have slightly or dramatically different appearances? Unlike in common human pose estimation benchmarks, here the setting is that datasets have many (annotated) poses per individual (>200) but only a few individuals ($\approx$10).

\begin{figure*}
    \begin{center}
    \includegraphics[width=\textwidth]{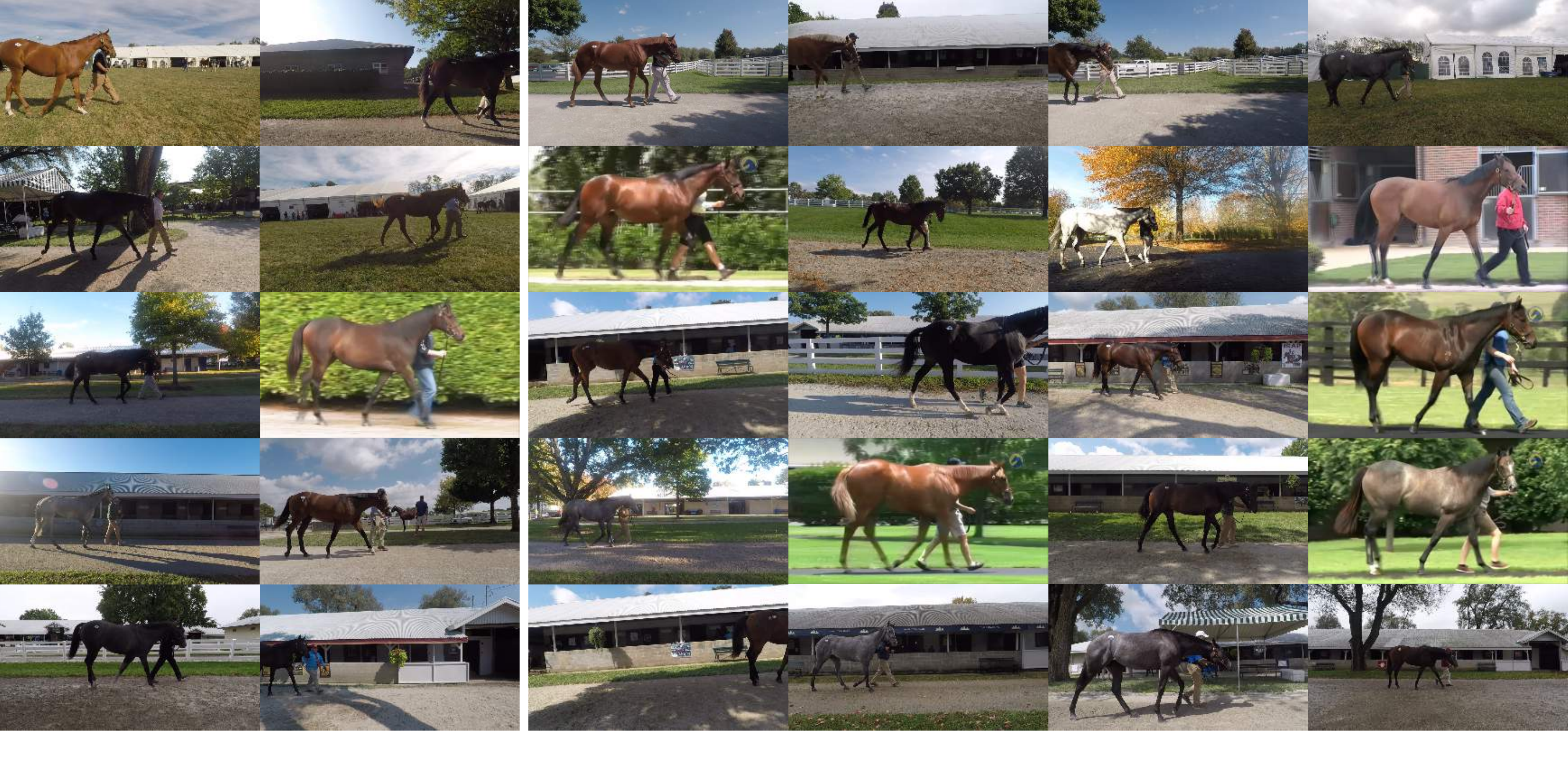}
    \end{center}
    \vspace{-7pt}
    \caption{{\bf Horse Dataset:} Example frames for each Thoroughbred horse in the dataset. The videos vary in horse color, background, lighting conditions, and relative horse size. The sunlight variation between each video added to the complexity of the learning challenge, as well as the handlers often wearing horse-leg-colored clothing. Some horses were in direct sunlight while others had the light behind them, and others were walking into and out of shadows, which was particularly problematic with a dataset dominated by dark colored coats. To illustrate the Horse-10 task we arranged the horses according to one split: the ten leftmost horses were used for train/test within-domain, and the rest are the out-of-domain held out horses.}
    \label{fig:horsedata}
\end{figure*}

To allow the field to tackle this challenge, we developed a novel benchmark comprising 30 diverse Thoroughbred horses, for which 22 body parts were labeled by an expert in \num{8114} frames (Dataset available at \url{http://horse10.deeplabcut.org}). Horses have various coat colors and the ``in-the-wild'' aspect of the collected data at various Thoroughbred farms added additional complexity. With this dataset we could directly test the effect of pretraining on out-of-domain data. Here we report two key insights: (1) ImageNet performance predicts generalization for both within domain and on out-of-domain data for pose estimation; (2) While we confirm that task-training can catch up with fine-tuning pretrained models given sufficiently large training sets~\cite{he2018rethinking}, we show this is not the case for out-of-domain data (Figure~\ref{fig:highlight}). Thus, transfer learning improves robustness and generalization. Furthermore, we contrast the domain shift inherent in this dataset with domain shift induced by common image corruptions~\cite{hendrycks2019benchmarking, michaelis2019dragon}, and we find pretraining on ImageNet also improves robustness against those corruptions.

\section{Related Work}

\subsection{Pose and keypoint estimation}

Typical human pose estimation benchmarks, such as MPII pose and COCO~\cite{lin2014microsoft,andriluka20142d,andriluka2018posetrack} contain many different individuals ($>10k$) in different contexts, but only very few example postures per individual. Along similar lines, but for animals, Cao et al. created a dataset comprising a few thousand images for five domestic animals with one pose per individual~\cite{cao2019cross}. There are also papers for facial keypoints in horses~\cite{rashid2017interspecies} and sheep~\cite{yang2016human, hewitt2019pose} and recently a large scale dataset featuring 21.9k faces from 334 diverse species was introduced~\cite{khan2020animalweb}. Our work adds a dataset comprising multiple different postures per individual (>200) and comprising 30 diverse race horses, for which 22 body parts were labeled by an expert in \num{8114} frames. This pose estimation dataset allowed us to address within and out of domain generalization. Our dataset could be important for further testing and developing recent work for domain adapation in animal pose estimation on a real-world dataset~\cite{li2020deformation, sanakoyeu2020transferring, mu2020learning}.

\begin{figure*}
    \begin{center}
    \includegraphics[width=\textwidth]{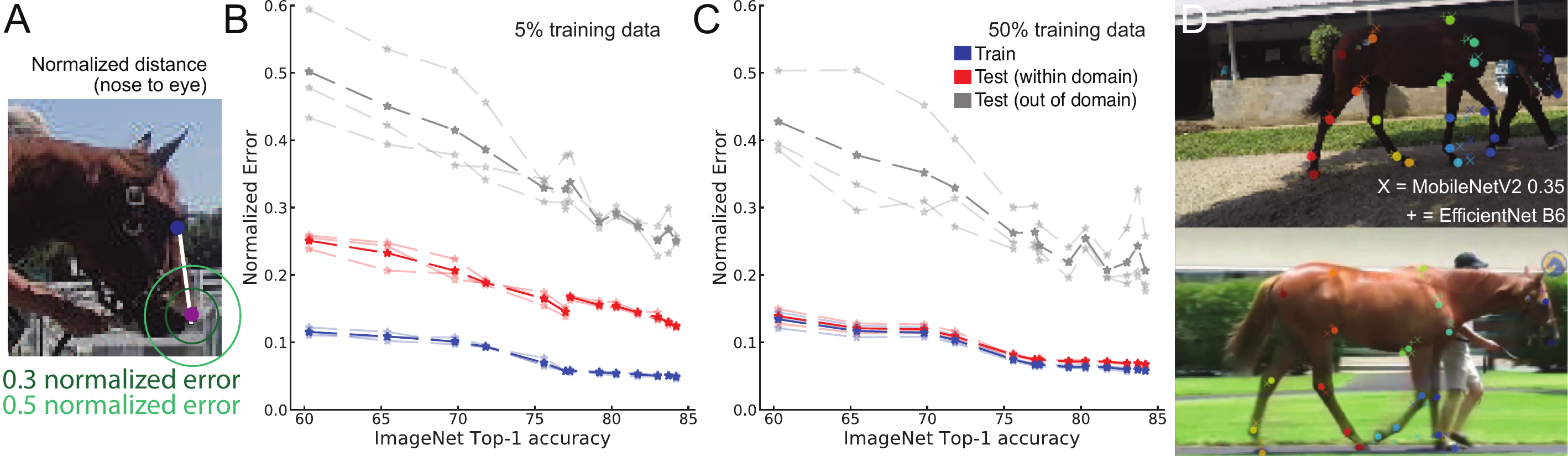}
    \end{center}
    \caption{ {\bf Transfer Learning boosts performance, especially on out-of-domain data.} {\bf A:} Illustration of normalized error metric, i.e. measured as a fraction of the distance from nose to eye (which is approximately \SI{30}{\centi\metre} on a horse). {\bf B:} Normalized Error vs. Network performance as ranked by the Top \SI{1}{\percent} accuracy on ImageNet (order by increasing ImageNet performance: MobileNetV2-0.35, MobileNetV2-0.5, MobileNetV2-0.75, MobileNetV2-1.0, ResNet-50, ResNet-101, EfficientNets B0 to B6). The faint lines indicate data for the three  splits. Test data is in red, train is blue, grey is out-of-domain data {\bf C:} Same as B but with \SI{50}{\percent} training fraction. {\bf D:} Example frames with human annotated body parts vs. predicted body parts for MobileNetV2-0.35 and EfficientNet-B6 architectures with ImageNet pretraining on out-of-domain horses.}
    \label{fig:TLresults}
    \vspace{-15pt}
\end{figure*}

\subsection{Transfer learning}

Transfer learning has become accepted wisdom: fine-tuning pretrained weights of large scale models yields best results~\cite{donahue2014decaf, Yosinski2014, kummerer2016deepgaze, mathis2018deeplabcut,li2019analysis, zhuang2019comprehensive}. He et al. nudged the field to rethink this accepted wisdom by demonstrating that for various tasks, directly training on the task-data can match performance~\cite{he2018rethinking}. We confirm this result, but show that on held-out individuals (``out-of-domain'') this is not the case. Raghu et al. showed that for target medical tasks (with little similarity to ImageNet) transfer learning offers little benefit over lightweight architectures~\cite{raghu2019transfusion}. Kornblith et al. showed for many object recognition tasks, that better ImageNet performance leads to better performance on these other benchmarks~\cite{kornblith2019better}. We show that this is also true for pose-estimation both for within-domain and out-of-domain data (on different horses, and for different species) as well as for corruption resilience.

What is the limit of transfer learning? Would ever larger data sets give better generalization? Interestingly, it appears to strongly depend on what task the network was pretrained on. Recent work by Mahajan et al. showed that pretraining for large-scale hashtag predictions on Instagram data (\num{3.5} billion images) improves classification, while at the same time possibly harming localization performance for tasks like object detection, instance segmentation, and keypoint detection~\cite{mahajan2018exploring}. This highlights the importance of the task, rather than the sheer size as a crucial factor. Further corroborating this insight, Li et al. showed that pretraining on large-scale object detection task can improve performance for tasks that require fine, spatial information like segmentation~\cite{li2019analysis}. Thus, one interesting future direction to boost robustness could be to utilize networks pretrained on OpenImages, which contains bounding boxes for 15 million instances and close to 2 million images~\cite{kuznetsova2018open}. 

\subsection{Robustness}

Studying robustness to common image corruptions based on benchmarks such as ImageNet-C~\cite{hendrycks2019benchmarking, michaelis2019dragon, Schneider2020inc} is a fruitful avenue for making deep learning more robust. Apart from evaluating our pose estimation algorithms on novel horses (domain-shift), we also investigate the robustness with respect to image corruptions. Hendrycks et al. study robustness to out-of distribution data on CIFAR 10, CIFAR 100 and TinyImageNet (but not pose estimation). The authors report that pretraining is important for adversarial robustness~\cite{Hendrycks2019}. Shah et al. found that pose estimation algorithms are highly robust against adversarial attacks~\cite{shah2019robustness}, but neither directly test out-of-domain robustness on different individuals, nor robustness to common image corruptions as we do in this study.

\section{Data and Methods} 
\subsection{Datasets and evaluation metrics} 

We developed a novel horse data set comprising \num{8114} frames across \num{30} different horses captured for \num{4}-\num{10} seconds with a GoPro camera (Resolution: $1920 \times 1080$, Frame Rate: $60$ FPS), which we call Horse-30 (Figure~\ref{fig:horsedata}). We downsampled the frames by a factor of $15\%$ to speed-up the benchmarking process ($288 \times 162$ pixels; one video was downsampled to $30\%$). We annotated 22 previously established anatomical landmarks for equines~\cite{magnusson1990studies,anderson2004longitudinal}. The following 22 body parts were labeled in \num{8114} frames: Nose, Eye, Nearknee, Nearfrontfetlock, Nearfrontfoot, 
Offknee, Offfrontfetlock, Offfrontfoot, 
Shoulder, Midshoulder, Elbow, Girth, Wither, 
Nearhindhock, Nearhindfetlock, Nearhindfoot, 
Hip, Stifle, Offhindhock, Offhindfetlock, Offhindfoot,
Ischium.
We used the DeepLabCut 2.0 toolbox~\cite{Nath2019} for labeling. We created 3 splits that contain 10 randomly selected training horses each (referred to as Horse-10). For each training set we took a subset of \SI{5}{\percent} ($\approx$\num{160} frames), and \SI{50}{\percent} ($\approx$ \num{1470} frames) of the frames for training, and then evaluated the performance on the training, test, and unseen (defined as ``out-of-domain'') horses (i.e. the other horses that were not in the given split of Horse-10). As the horses could vary dramatically in size across frames, due to the ``in-the-wild'' variation in distance from the camera, we normalized the raw pixel errors by the eye-to-nose distance and report the fraction of this distance (normalized error) as well as percent correct keypoint metric~\cite{andriluka20142d}; we used a matching threshold of \SI{30}{\percent} of the head segment length (nose to eye per horse; see Figure~\ref{fig:TLresults}A).

For the generalization experiments, we also tested on the Animal Pose dataset~\cite{cao2019cross} to test the  generality of our findings (Figure~\ref{fig:AnimalPose}). We extracted all single animal images from this dataset, giving us \num{1091} cat, \num{1176} dog, \num{486} horse, \num{237} cow, and \num{214} sheep images. To note, we corrected errors in ground truth labels for the dog's (in about \SI{10}{\percent} of frames). Because nearly all images in this dataset are twice the size of the Horse-10 data, we also downsampled the images by a factor of 2 before training and testing. Given the lack of a consistent eye-to-nose distance across the dataset due to the varying orientations, we normalized as follows: the raw pixel errors were normalized by the square root of the bounding box area for each individual. For training the various architectures, the best schedules from cross validation on Horse-10 were used (see Section 3.2).

We also applied common image corruptions~\cite{hendrycks2019benchmarking} to the Horse-10 dataset, yielding a variant of the benchmark which we refer to as Horse-C. Horse-C images are corrupted with 15 forms of digital transforms, blurring filters, point-wise noise or simulated weather conditions. All conditions are applied following the evaluation protocol and implementation by Michaelis et al.~\cite{michaelis2019dragon}. In total, we arrived at 75 variants of the dataset (15 different corruptions at 5 different severities), yielding over 600k images.

\subsection{Architectures and Training Parameters} 

We utilized the pose estimation toolbox called DeepLabCut~\cite{mathis2018deeplabcut, Nath2019, insafutdinov2016deepercut}, and added MobileNetV2~\cite{sandler2018mobilenetv2} and EfficientNet backbones~\cite{tan2019efficientnet} to the ResNets~\cite{He_2016_CVPR} that were present, as well as adding imgaug for data augmentation~\cite{imgaug}. The TensorFlow~\cite{abadi2016tensorflow}-based network architectures could be easily exchanged while keeping data loading, training, and evaluation consistent. The feature detectors in DeepLabCut consist of a backbone followed by deconvolutional layers to predict pose scoremaps and location refinement maps (offsets), which can then be used for predicting the pose while also providing a confidence score. As previously, for the ResNet backbones we utilize an output stride of 16 and then upsample the filter banks with deconvolutions by a factor of two to predict the heatmaps and location-refinement at \num{1 / 8}th of the original image size scale~\cite{insafutdinov2016deepercut,mathis2018deeplabcut}. For MobileNetV2~\cite{sandler2018mobilenetv2}, we configured the output-stride as $16$ (by changing the last stride $2$ convolution to stride $1$). 

We utilized four variants of MobileNetV2 with different expansion ratios (\num{0.35}, \num{0.5}, \num{0.75} and \num{1}) as this ratio modulates the ImageNet accuracy from \SI{60.3}{\percent} to \SI{71.8}{\percent}, and pretrained models on ImageNet from TensorFlow~\cite{abadi2016tensorflow, sandler2018mobilenetv2}.

The baseline EfficientNet model was designed by Tan et al.~\cite{tan2019efficientnet} through a neural architecture search to optimize for accuracy and inverse FLOPS.
From B0 to B6, compound scaling is used to increase the width, depth, and resolution of the network, which directly corresponds to an increase in ImageNet performance~\cite{tan2019efficientnet}.
We used the AutoAugment pretrained checkpoints from TensorFlow as well as adapted the EfficientNet's output-stride to $16$ (by changing the (otherwise) last stride $2$ convolution to stride $1$). 

The training loss is defined as the cross entropy loss for the scoremaps and the location refinement error via a Huber loss with weight $0.05$~\cite{mathis2018deeplabcut}. The loss is minimized via ADAM with batch size 8~\cite{kingma2014adam}. For training, a cosine learning rate schedule, as in ~\cite{kornblith2019better} with ADAM optimizer and batchsize 8 was used; we also performed augmentation, using imgaug~\cite{imgaug}, with random cropping and rotations. Initial learning rates and decay target points were cross-validated for MobileNetV2 $0.35$ and $1.0$, ResNet-$50$, EfficientNet B0, B3, and B5 for the pretrained and from scratch models (see Supplementary Material). For each model that was not cross validated (MobileNetV2 $0.5$ and $0.75$, ResNet-$101$, EfficientNet B1, B2, B4, B6), the best performing training parameters from the most similar cross validated model was used (i.e. the cross validated EfficientNet-B0 schedule was used for EfficientNet-B1; see Supplementary Material). For MobileNetV2s, we trained the batch normalization layers too (this had little effect on task performance for MobileNetV2-$0.35$). Pretrained models were trained for 30k iterations (as they converged), while models from scratch were trained for 180k iterations. From scratch variants of the architectures used He-initialization~\cite{he2015initialization}, while all pretrained networks were initialized from their ImageNet trained weights.

\subsection{Cross Validation of Learning Schedules}

To fairly compare the pose estimation networks with different backbones, we cross-validated the learning schedules. For models with pretraining and from scratch, we cross validated the cosine learning rate schedules by performing a grid search of potential initial learning rates and decay targets to optimize their performance on out of domain data. Given that our main result is that while task-training can catch up with fine-tuning pretrained models given sufficiently large training sets on within-domain-data (consistent with~\cite{he2018rethinking}), we will show that this is not the case for out-of-domain data. Thus, in order to give models trained from scratch the best shot, we  optimized the performance on out of domain data. Tables in the Supplementary Material describe the various initial learning rates explored during cross validation as well as the best learning schedules for each model.
 
\subsection{Similarity Analysis}

To elucidate the differences between pretrained models and models trained from scratch, we analyze the representational similarity between the variants. We use linear centered kernel alignment (CKA)~\cite{kornblith2019similarity} to compare the image representations at various depths in the backbone networks. The results were aggregated with respect to the similarity of representations of within domain images versus out of domain images, and averaged over the three shuffles.

\section{Results} 

To test within and out-of-domain performance we created a new dataset of $30$ different Thoroughbreds that are led by different humans, resulting in a dataset of \num{8114} images with \num{22} labeled body parts each. These videos differ strongly in horse appearance, context, and background  (Figure~\ref{fig:horsedata}). Thus, this dataset is ideal for testing robustness and out-of-sample generalization. We created 3 splits containing 10 random horses each, and then varied the amount of training data from these 10 horses (referred to as Horse-10, see Methods). As the horses could vary dramatically in size across frames, due to the ``in-the-wild'' variation in distance from the camera, we used a normalized pixel error; i.e. we normalized the raw pixel errors by the eye-to-nose distance and report the fraction within this distance (see Methods).

\begin{table}[t]
\caption{average PCK@0.3 (\%) }
\label{aPCK-table}
\begin{center}
\begin{sc}
\footnotesize
\begin{tabular}{lcccr}
\toprule
Models & Within Domain & Out-of-D. \\
\midrule
MobileNetV2-0.35 & 95.2 & 63.5\\
MobileNetV2-0.5 & 97.1 & 70.4\\
MobileNetV2-0.75 & 97.8 & 73.0\\
MobileNetV2-1 & 98.8 & 77.6\\
ResNet-50 & 99.8 & 81.3 \\
ResNet-101  & 99.9 & 84.3\\
EfficientNet-B0 & 99.9 & 81.6 \\
EfficientNet-B1 & 99.9 & 84.5 \\
EfficientNet-B2 & 99.9 & 84.3 \\
EfficientNet-B3 & 99.9 & 86.6 \\
EfficientNet-B4 & 99.9 & 86.9\\
EfficientNet-B5 & 99.9 & 87.7\\
EfficientNet-B6 & 99.9 & 88.4\\
\bottomrule
\end{tabular}
\end{sc}
\end{center}
\vspace{-25pt}
\end{table}

\begin{table*}[ht]
\centering
\caption{PCK@0.3 (\%) for several bodyparts and architectures on within domain horses (FF=front foot; HF = Hind foot; HH = Hind Hock).}
\label{PCK-tableperbptwithin} \footnotesize
\begin{tabular}{lcccccccccccc}
\toprule
{} &    Nose &    Eye &  Shoulder &  Wither &  Elbow &  NearFF &  OffFF &    Hip &  NearHH &  NearHF &  OffHF  \\
\midrule
MobileNetV2 0.35 &   90.7 &   94.1 &      97.6 &    96.9 &   96.7 &           92.3 &          93.7 &   96.4 &          94.1 &          94.2 &         92.5  \\
MobileNetV2 0.5  &   94.1 &   96.1 &      99.2 &    98.3 &   98.0 &           93.8 &          95.4 &   96.7 &          97.2 &          97.2 &         97.0  \\
MobileNetV2 0.75 &   96.0 &   97.5 &      99.2 &    98.0 &   99.0 &           96.6 &          96.8 &   98.8 &          97.6 &          98.0 &         97.4  \\
MobileNetV2 1.0  &   97.7 &   98.8 &      99.7 &    99.1 &   99.0 &           97.6 &          97.3 &   99.4 &          98.4 &          98.5 &         98.9  \\
ResNet 50        &   99.9 &  100.0 &      99.8 &    99.9 &   99.8 &           99.8 &          99.6 &   99.9 &          99.9 &          99.6 &         99.8  \\
ResNet 101       &   99.9 &  100.0 &      99.9 &    99.8 &   99.9 &           99.8 &          99.7 &   99.8 &          99.9 &          99.7 &         99.9  \\
EfficientNet-B0  &   99.7 &   99.9 &     100.0 &    99.9 &  100.0 &           99.6 &          99.5 &  100.0 &          99.9 &          99.7 &         99.7  \\
EfficientNet-B1  &   99.8 &   99.9 &     100.0 &    99.8 &   99.9 &           99.5 &          99.8 &  100.0 &          99.8 &          99.8 &         99.8  \\
EfficientNet-B2  &   99.9 &   99.9 &     100.0 &    99.9 &  100.0 &           99.8 &          99.7 &   99.9 &          99.8 &          99.7 &         99.7  \\
EfficientNet-B3  &   99.9 &   99.9 &      99.9 &    99.9 &   99.9 &           99.7 &          99.6 &   99.7 &          99.8 &          99.6 &         99.9  \\
EfficientNet-B4  &  100.0 &  100.0 &      99.9 &    99.8 &   99.9 &           99.6 &          99.7 &   99.9 &          99.7 &          99.8 &         99.8  \\
EfficientNet-B5  &   99.9 &   99.9 &     100.0 &    99.9 &  100.0 &           99.7 &          99.8 &   99.6 &          99.8 &          99.8 &         99.9  \\
EfficientNet-B6  &   99.9 &   99.9 &      99.9 &    99.8 &  100.0 &           99.8 &          99.9 &   99.8 &          99.8 &          99.7 &         99.8\\ 
\bottomrule
\end{tabular}
\end{table*}

\begin{table*}[ht]
\caption{PCK@0.3 (\%) for several bodyparts and architectures on out-of-domain horses (FF=front foot; HF = Hind foot; HH = Hind Hock).}
\label{PCK-tableperbptacross} \footnotesize
\centering
\begin{tabular}{lcccccccccccc}
\toprule
{} &    Nose &    Eye &  Shoulder &  Wither &  Elbow &  NearFF &  OffFF &    Hip &  NearHH &  NearHF &  OffHF  \\
\midrule
MobileNetV2 0.35 &  45.6 &  53.1 &      65.5 &    68.0 &   69.1 &           56.4 &          57.6 &  65.9 &          65.9 &          60.5 &         62.5  \\
MobileNetV2 0.5  &  52.7 &  61.0 &      76.7 &    69.7 &   78.3 &           62.9 &          65.4 &  73.6 &          70.8 &          68.1 &         69.7  \\
MobileNetV2 0.75 &  54.2 &  65.6 &      78.3 &    73.2 &   80.5 &           67.3 &          68.9 &  80.0 &          74.1 &          70.5 &         70.2  \\
MobileNetV2 1.0  &  59.0 &  67.2 &      83.8 &    79.7 &   84.0 &           70.1 &          72.1 &  82.0 &          79.9 &          76.0 &         76.7  \\
ResNet 50        &  68.2 &  73.6 &      85.4 &    85.8 &   88.1 &           72.6 &          70.2 &  89.2 &          85.7 &          77.0 &         74.1  \\
ResNet 101       &  67.7 &  72.4 &      87.6 &    86.0 &   89.0 &           79.9 &          78.0 &  92.6 &          87.2 &          83.4 &         80.0  \\
EfficientNet-B0  &  60.3 &  62.5 &      84.9 &    84.6 &   87.2 &           77.0 &          75.4 &  86.7 &          86.7 &          79.6 &         79.4  \\
EfficientNet-B1  &  67.4 &  71.5 &      85.9 &    85.7 &   89.6 &           80.0 &          81.1 &  86.7 &          88.4 &          81.8 &         81.6  \\
EfficientNet-B2  &  68.7 &  74.8 &      84.5 &    85.2 &   89.2 &           79.7 &          80.9 &  88.1 &          88.0 &          82.3 &         81.7 \\
EfficientNet-B3  &  71.7 &  76.6 &      88.6 &    88.7 &   92.0 &           80.4 &          81.8 &  90.6 &          90.8 &          85.0 &         83.6  \\
EfficientNet-B4  &  71.1 &  75.8 &      88.1 &    87.4 &   91.8 &           83.3 &          82.9 &  90.8 &          90.3 &          86.7 &         85.5  \\
EfficientNet-B5  &  74.8 &  79.5 &      89.6 &    89.5 &   93.5 &           82.2 &          84.1 &  91.8 &          90.9 &          86.6 &         85.2  \\
EfficientNet-B6  &  74.7 &  79.7 &      90.3 &    89.8 &   92.8 &           83.6 &          84.4 &  92.1 &          92.1 &          87.8 &         85.3  \\
\bottomrule
\end{tabular}

\end{table*}

\begin{figure}[h]
\includegraphics[width=\columnwidth]{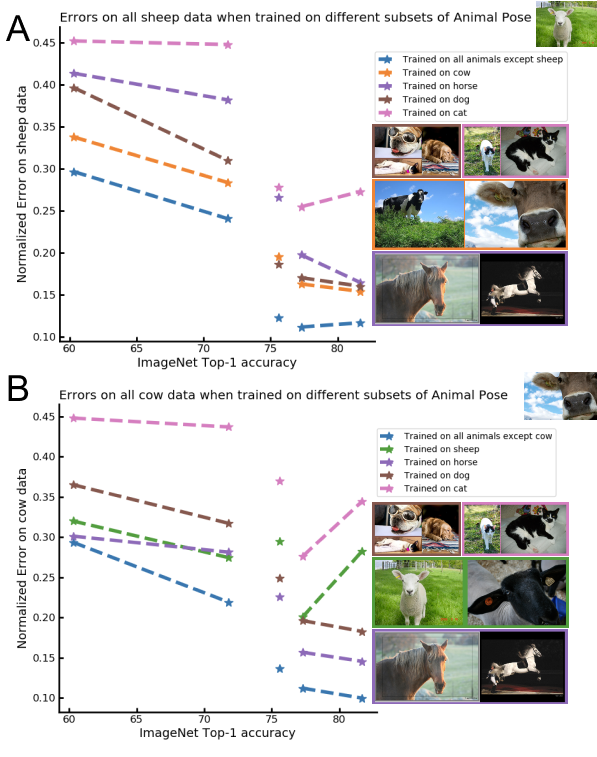}
\caption{ {\bf Generalization Across Species.} Normalized pose estimation error vs. ImageNet Top 1\% accuracy with different backbones (as in Figure~\ref{fig:highlight}), but for 10 additional out-of-domain tests.  Training on either a single species or four species while holding one species (either cow or sheep) out.}
\vspace{-10pt}
\label{fig:AnimalPose}
\end{figure}
\subsection{ImageNet performance vs task performance}

To probe the impact of ImageNet performance on pose estimation robustness, we selected modern convolutional architectures as backbones with a wide range of ImageNet performance (see Methods; 13 models spanning from $60\%$ to $84\%$ ImageNet performance). To fairly compare the MobileNetV2, ResNet and EfficientNet backbones, we cross validated the learning schedules for each model (see Methods).  In total, we found that all ImageNet-pretrained architectures exhibited strong performance on Horse-10 within domain, i.e. low average errors, and high average percent correct key points (aPCK; Figure~\ref{fig:TLresults}, Table~\ref{aPCK-table}). Performance on Horse-10 within domain also closely matched performance on Horse-30 (see Supplementary Material). Next, we directly compared the ImageNet performance to their respective performance on this pose estimation task. We found Top-$1\%$ ImageNet accuracy correlates with pose estimation test error (linear fit for test: slope $-0.33\%$, $R^2=0.93$, $p=\num{1.4e-7}$; Figure~\ref{fig:TLresults}). Results for different bodyparts are displayed in Table~\ref{PCK-tableperbptwithin}.

\subsection{Generalization to novel horses}

Next, we evaluated the performance of the networks on different horses in different contexts, i.e. out-of-domain horses (Figures~\ref{fig:TLresults}A-C). Most strikingly, on out-of-domain horses, the relationship between ImageNet performance and performance on Horse-10 was even stronger. This can be quantified by comparing the linear regression slope for out-of-domain test data: $-0.93\%$ pose-estimation improvement per percentage point of ImageNet performance, $R^2=\num{0.93}$, $p=\num{9e-8}$ vs. within-domain test data $-0.33\%$, $R^2=0.93$, $p=\num{1.4e-7}$ (for $50\%$ training data). Results for several different bodyparts of the full 22 are displayed in Table~\ref{PCK-tableperbptacross}, highlighting that better models also generalized better in a bodypart specific way. In other words, {\it less} powerful models overfit more on the training data.

\subsection{Generalization across species}

Does the improved generalization to novel individuals also hold for a more difficult out-of-domain generalization, namely, across species?
Thus, we turned to a pose-estimation dataset comprising multiple species. We evaluated the performance of the various architectures on the Animal Pose dataset from Cao et. al~\cite{cao2019cross}. Here, images and poses of horses, dogs, sheep, cats, and cows allow us to test performance across animal classes. Using ImageNet pretraining and the cross validated schedules from our Horse-10 experiments, we trained on individual animal classes or multiple animal classes (holding out sheep/cows) and examined how the architectures generalized to sheep/cows, respectively (Figure~\ref{fig:AnimalPose}). For both cows and sheep, better ImageNet architectures, trained on the pose data of other animal classes performed better, in most cases. 
We mused that this improved generalization could be a consequence of the ImageNet pretraining or the architectures themselves. Therefore, we turned to Horse-10 and trained the different architectures directly on horse pose estimation from scratch.

\begin{figure*}
    \begin{center}
    \includegraphics[width=.9\textwidth]{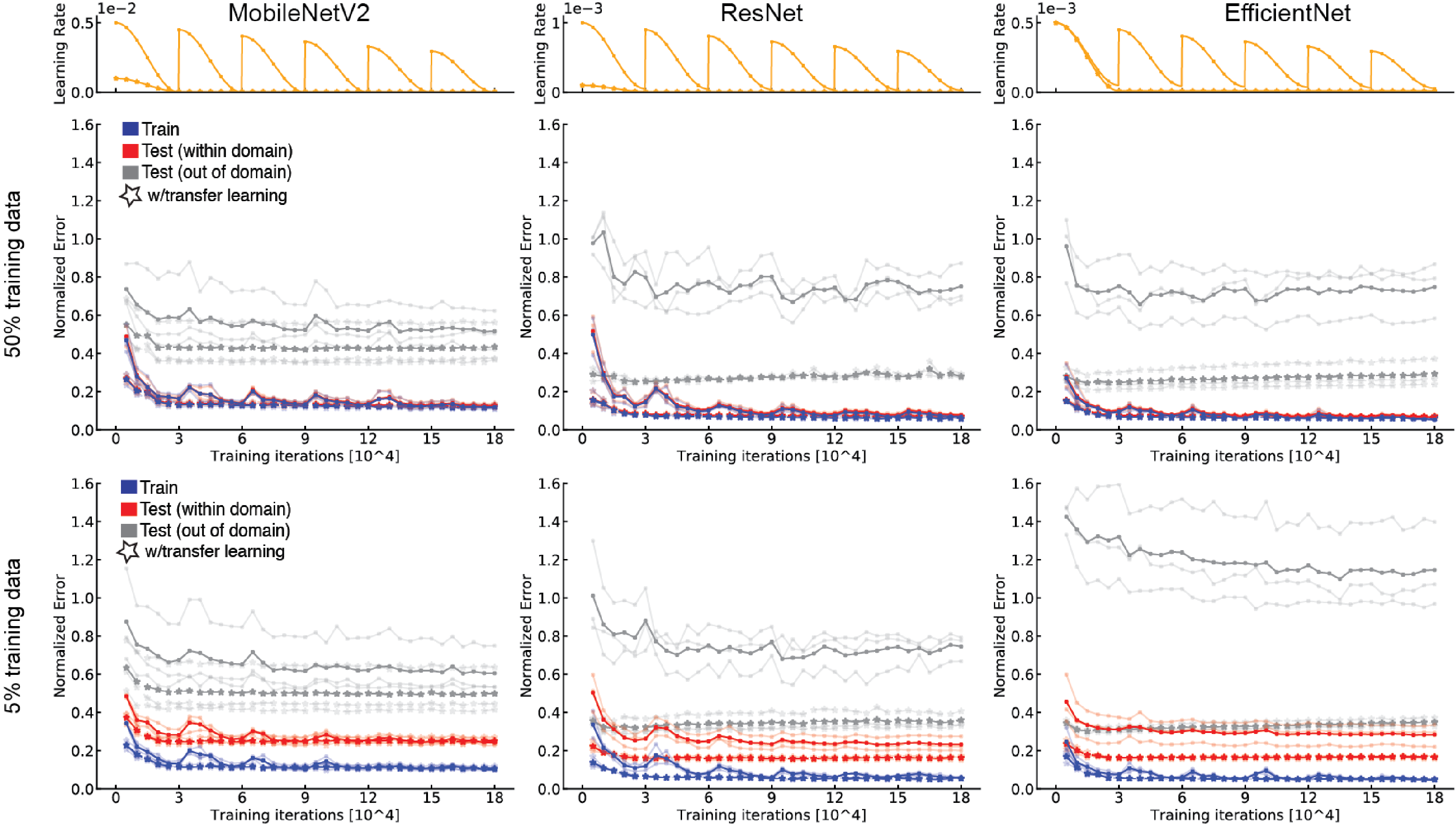}
    \end{center}
    \vspace{-7pt}
    \caption{{\bf Training randomly initialized networks longer cannot rescue out-of-domain performance.} {\bf Top Row:} Best performing (cross-validated) learning schedules used for training. {\bf Middle:} Normalized error vs. training iterations for MobileNetV2-0.35, ResNet-50 and EfficientNet-B0 {\it using $50\%$ of the training data}. Test errors when training from scratch (solid lines) closely match the transfer learning (dashed lines) performance after many iterations. Crucially, out-of-domain testing does not approach performance for pretrained network (stars). {\bf Bottom Row:} Same as Middle but {\it using $5\%$ of the training data}; note, however, for just $5\%$ training data, the test errors do not approach the test error of pretrained models for larger models.}
    \label{fig:FromScratchLonger}
    \vspace{-10pt}
\end{figure*}

\subsection{Task-based training from scratch} 

To assess the impact of ImageNet pretraining we also trained several architectures from scratch. Thereby we could directly test if the increased slope for out-of-domain performance across networks was merely a result of more powerful network architectures. He et al. demonstrated that training Mask R-CNN with ResNet backbones directly on the COCO object detection, instance segmentation and key point detection tasks, {\it catches-up} with the performance of ImageNet-pretrained variants if training for substantially more iterations than typical training schedules~\cite{he2018rethinking}. However, due to the nature of these tasks, they could not test this relationship on out-of-domain data. For fine-tuning from ImageNet pretrained models, we trained for $30k$ iterations (as the loss had flattened; see Figure~\ref{fig:FromScratchLonger}). First, we searched for best performing schedules for training from scratch while substantially increasing the training time ($6X$ longer). We found that cosine decay with restart was the best for out-of-domain performance (see Methods; Figure~\ref{fig:FromScratchLonger}).

Using this schedule, and consistent with He et al.~\cite{he2018rethinking}, we found that randomly initialized networks could closely match the performance of pretrained networks, given enough data and time (Figure~\ref{fig:FromScratchLonger}). As expected, for smaller training sets (\SI{5}{\percent} training data; 160 images), this was not the case (Figure~\ref{fig:FromScratchLonger}). While task-training could therefore match the performance of pretrained networks given enough training data, this was not the case for novel horses (out-of-domain data). The trained from-scratch networks never caught up and indeed plateaued early (Figure~\ref{fig:FromScratchLonger}; Figure~\ref{fig:highlight}). Quantitatively, we also found that for stronger networks (ResNets and EfficientNets) generalization was worse if trained from scratch (Figure~\ref{fig:highlight}). Interestingly that was not the case for the lightweight models, i.e. MobileNetV2s (cf.~\cite{raghu2019transfusion}).

\subsection{Network similarity analysis} 

We hypothesized that the differences in generalization are due to more invariant representations in networks with higher ImageNet-performance using Centered Kernel Alignment (CKA)~\cite{kornblith2019similarity}. We first verified that the representations change with task training (Supplementary Material Figure 1). 
We compared the representations of within-domain and out-of-domain images across networks trained from ImageNet vs. from scratch. We found that early blocks are similar for from scratch vs transfer learning for both sets of horses. In later layers, the representations diverge, but comparisons between within-domain and out of domain trends were inconclusive as to why e.g., EfficientNets generalize better (Supplementary Material Figure 2).

\subsection{Horse-C: Robustness to image corruptions} 

\begin{figure*}[ht]
    \centering
    \includegraphics[width=\textwidth]{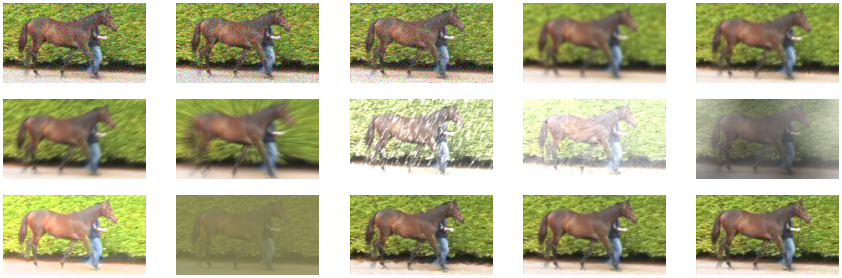}
    \caption{%
    \textbf{Measuring the impact of common image corruptions on pose estimation (Horse-C)}: We adapt the image corruptions considered by Hendrycks et al. and contrast the impact of common image corruptions with that of out of domain evaluation. 
    }
    \label{fig:HorseC}
\end{figure*}

\begin{figure}[h]
    \centering
    \includegraphics[height=.56\columnwidth]{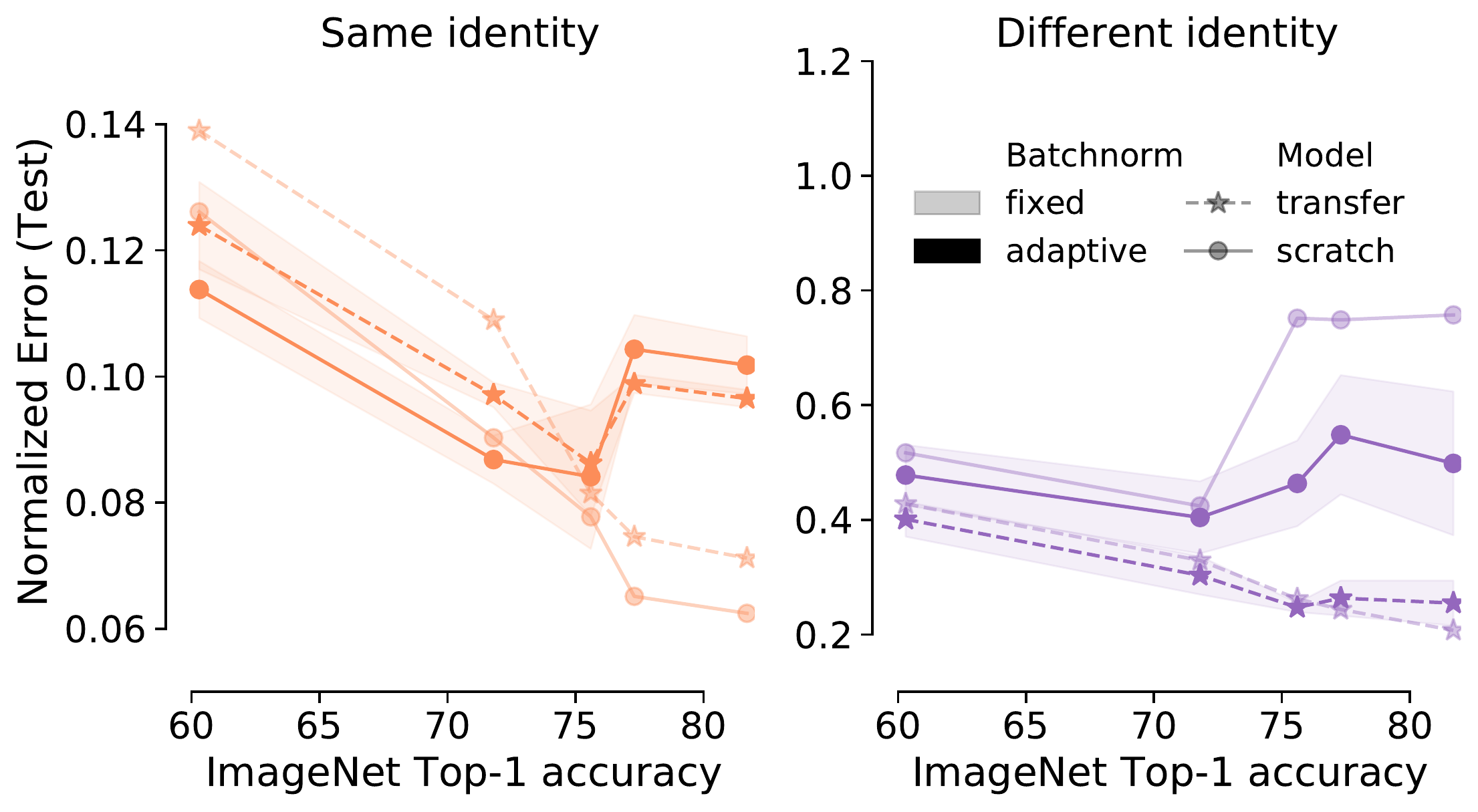}
    \caption{\textbf{Impact of test time normalization.}  Models trained with adaptive BN layers slightly outperform our baseline models for the MobileNetV2 and ResNet architecture out-of-domain evaluation. Lines with alpha transparency represent fixed models (vs. adapted). We show mean $\pm$ SEM computed across 3 data splits.
    }
    \label{fig:comparison-bn-robustness}
    \vspace{-.5em}
\end{figure}

\begin{figure}[h]
    \centering
    \includegraphics[height=.56\columnwidth]{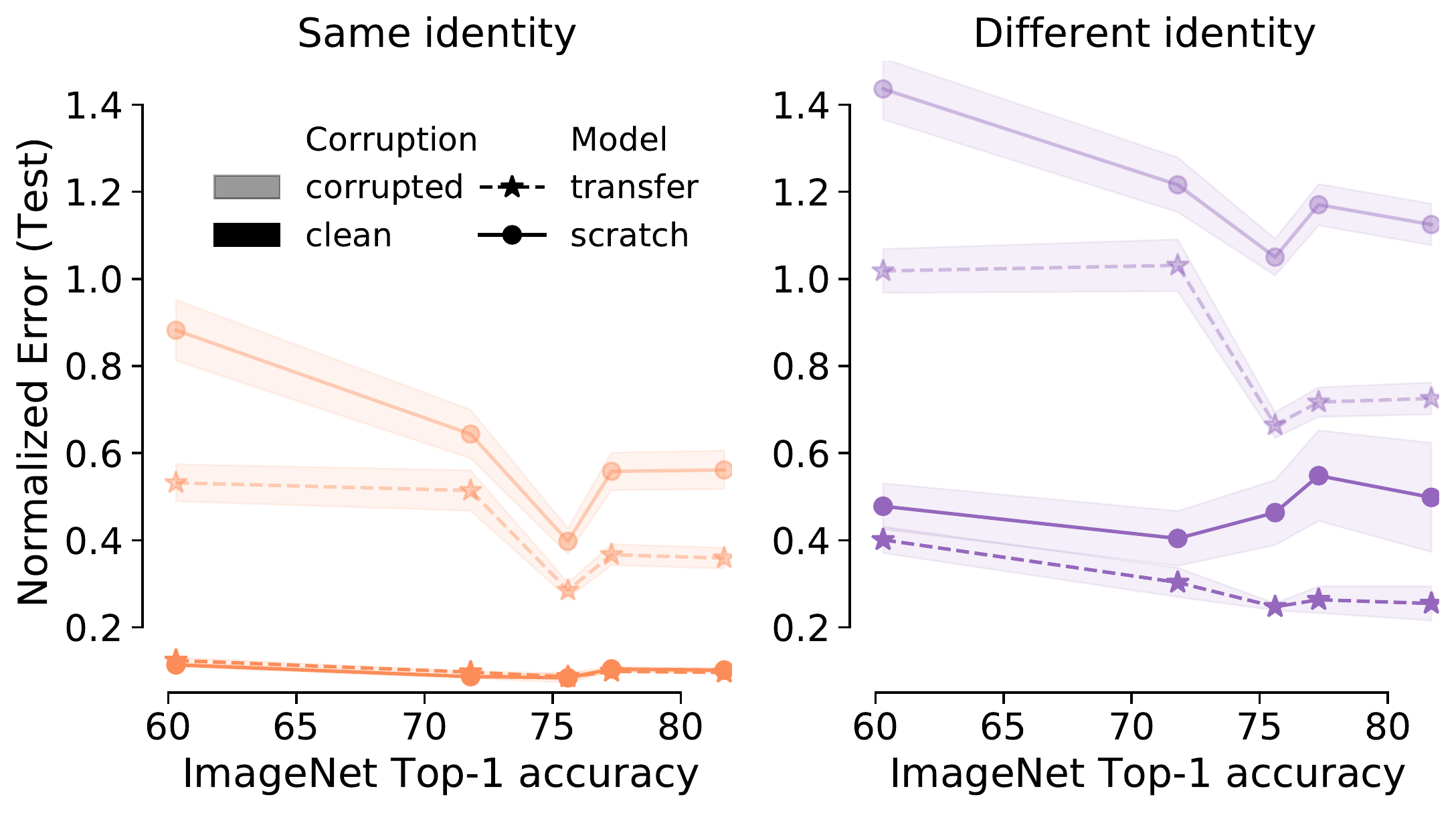}
    \caption{\textbf{Impact of distribution shift introduced by horse identities and common corruptions.} We tested within identity (i.e., equivalent to within-domain in Horse-10 (left), or out-of-domain identity (right). Lines with alpha transparency represent corrupted images, whereas solid is the original (clean) image. We show mean $\pm$ SEM across 3 splits for clean images, and across 3 splits, 15 corruptions and 5 severities for corrupted images.
    }
    \label{fig:comparison-horsec-robustness}
\end{figure}

To elucidate the difficulty of the Horse-10 benchmark, we more broadly evaluate pose estimation performance under different forms of domain shift (Figure~\ref{fig:HorseC}). Recently, Schneider, Rusak et al. demonstrated that simple unsupervised domain adaptation methods can greatly enhance the performance on corruption robustness benchmarks~\cite{Schneider2020inc}. We therefore settled on a full adaptation evaluation protocol: We re-trained MobileNetV2 0.35 and 1.0, ResNet50, as well as EfficientNet B0 and B3 with batch normalization enabled. During evaluation we then re-computed separate batch norm statistics for each horse and corruption type.

We use batch norm adaptation \cite{Schneider2020inc} during our evaluation on Horse-C. On clean out-of-domain data, we see improvements for MobileNetV2s and ResNets when using pre-trained networks, and for all models when training models from scratch (Figure~\ref{fig:comparison-bn-robustness}).
On common corruptions, utilizing adaptation is crucial to final performance (see full results in Supplementary Material). 
In the batch norm adapted models, we compared four test conditions comprised of within-domain and out-of domain for both original (clean) and corrupted images (Figure~\ref{fig:HorseC}). First, we find that even with batch norm adapted models, Horse-C is as hard as Horse-10; namely performance is significantly affected on corrupted data (Figure~\ref{fig:comparison-horsec-robustness}). Secondly, we find the corruption plus ``out-of-domain'' identity, is even harder---the performance degradation induced by different horse identities is on the same order of magnitude as the mean error induced on the corrupted dataset. Finally, and consistent with our other results, we found a performance gain by using pretrained networks (Figure~\ref{fig:comparison-horsec-robustness}).

\section{Discussion and conclusions} 

We developed a novel pose estimation benchmark for out-of-domain robustness (Horse-10), and for testing common image corruptions on pose estimation (Horse-C). The data and benchmarks are available at \url{http://horse10.deeplabcut.org}. Furthermore, we report two key findings: (1) pretrained-ImageNet networks offer known advantages: shorter training times, and less data requirements, as well as a novel advantage:  robustness on out-of-domain data, and (2) pretained networks that have higher ImageNet performance lead to better generalization. Collectively, this sheds a new light on the inductive biases of ``better ImageNet architectures'' for visual tasks to be particularly beneficial for robustness.

We introduced novel DeepLabCut model variants, part of \url{https://github.com/DeepLabCut/DeepLabCut}, that can achieve high accuracy, but with higher inference speed (up to double) than the original ResNet backbone (see Supplementary Material for an inference speed benchmark). 

In summary, transfer learning offers multiple advantages. Not only does pretraining networks on ImageNet allow for using smaller datasets and shorter training time (Figure~\ref{fig:FromScratchLonger}), it also strongly improves robustness and generalization, especially for more powerful, over-parameterized models. In fact, we found a strong correlation between generalization and ImageNet accuracy (Figure~\ref{fig:TLresults}). While we found a significant advantage ($>$2X boost) of using pretrained networks vs. from scratch for out-of-domain robustness, there is still a 3-fold difference in performance between within domain and out of domain (Figure~\ref{fig:highlight}). We believe that this work demonstrates that transfer learning approaches are powerful to build robust architectures, and are particularly important for further developing performance improvements on real-world datasets, such as Horse-10 and derived benchmarks such as Horse-C. Furthermore, by sharing our animal pose robustness benchmark dataset, we also believe that the community can collectively work towards closing the gap.

\subsection*{Acknowledgements}

Funding was provided by a Rowland Fellowship (MWM), CZI EOSS Grant (MWM \& AM), the Bertarelli Foundation (MWM) and
the German Federal Ministry of Education and Research (BMBF) through the Tübingen AI Center (FKZ: 01IS18039A; StS \& MB).
St.S. thanks the International Max Planck Research School for Intelligent Systems and acknowledges his membership in the European Laboratory for Learning and Intelligent Systems (ELLIS) PhD program.
We thank Maxime Vidal for ground truth corrections to the Animal Pose dataset.

{\small
\bibliographystyle{ieee_fullname}
\bibliography{references}
}

\newpage
\onecolumn
\appendix
\section{Additional information on the Horse-10 dataset}
The table lists the following statistics: labeled frames, scale (nose-to-eye distance in pixels), and whether a horse was within domain (w.d) or out-of-domain (o.o.d.) for each shuffle.
\begin{table*}[ht]
\begin{center}
\small
\begin{tabular}{lccccc}
\toprule
{} Horse Identifier &     samples &  nose-eye dist & shuffle 1 & shuffle 2 & shuffle 3 \\
\midrule
BrownHorseinShadow         &  308 &      22.3 &     o.o.d &     o.o.d &      w.d. \\
BrownHorseintoshadow       &  289 &      17.4 &     o.o.d &     o.o.d &     o.o.d \\
Brownhorselight            &  306 &      15.57 &     o.o.d &      w.d. &     o.o.d \\
Brownhorseoutofshadow      &  341 &      16.22 &     o.o.d &      w.d. &      w.d. \\
ChestnutHorseLight         &  318 &      35.55 &      w.d. &      w.d. &     o.o.d \\
Chestnuthorseongrass       &  376 &      12.9 &     o.o.d &      w.d. &      w.d. \\
GreyHorseLightandShadow    &  356 &      14.41 &      w.d. &      w.d. &     o.o.d \\
GreyHorseNoShadowBadLight  &  286 &      16.46 &      w.d. &     o.o.d &      w.d. \\
TwoHorsesinvideobothmoving &  181 &      13.84 &     o.o.d &     o.o.d &      w.d. \\
Twohorsesinvideoonemoving  &  252 &      16.51 &      w.d. &      w.d. &      w.d. \\
Sample1                    &  174 &      24.78 &     o.o.d &     o.o.d &     o.o.d \\
Sample2                    &  330 &      16.5 &     o.o.d &     o.o.d &     o.o.d \\
Sample3                    &  342 &      16.08 &     o.o.d &     o.o.d &     o.o.d \\
Sample4                    &  305 &      18.51 &     o.o.d &     o.o.d &      w.d. \\
Sample5                    &  295 &      16.89 &      w.d. &     o.o.d &     o.o.d \\
Sample6                    &  376 &      12.3 &     o.o.d &     o.o.d &     o.o.d \\
Sample7                    &  262 &      18.52 &      w.d. &     o.o.d &     o.o.d \\
Sample8                    &  388 &      12.5 &      w.d. &      w.d. &     o.o.d \\
Sample9                    &  359 &      12.43 &     o.o.d &     o.o.d &     o.o.d \\
Sample10                   &  235 &      25.18 &     o.o.d &     o.o.d &     o.o.d \\
Sample11                   &  256 &      19.16 &     o.o.d &      w.d. &     o.o.d \\
Sample12                   &  288 &      17.86 &      w.d. &     o.o.d &      w.d. \\
Sample13                   &  244 &      25.78 &      w.d. &      w.d. &      w.d. \\
Sample14                   &  168 &      25.55 &     o.o.d &     o.o.d &     o.o.d \\
Sample15                   &  154 &      26.53 &     o.o.d &     o.o.d &     o.o.d \\
Sample16                   &  212 &      15.43 &     o.o.d &     o.o.d &     o.o.d \\
Sample17                   &  240 &      10.04 &      w.d. &     o.o.d &     o.o.d \\
Sample18                   &  159 &      29.55 &     o.o.d &      w.d. &     o.o.d \\
Sample19                   &  134 &      13.44 &     o.o.d &     o.o.d &      w.d. \\
Sample20                   &  180 &      28.57 &     o.o.d &     o.o.d &     o.o.d \\
\bottomrule
mean                       &  270.47 &      18.89 &        &       &       \\
STD                        &   73.04 &       6.05 &      &        &        \\
\end{tabular}
\end{center}
\end{table*}

\subsection{Learning schedule cross validation}
\label{crossval}

Because of the extensive resources required to cross validate all models, we only underwent the search on MobileNetV2s 0.35 and 1.0, ResNet 50, and EfficientNets B0, B3, and B5 for the pretraining and from scratch variants. For all other models, the parameters from the most similar networks were used for training (i.e. EfficientNet-B1 used the parameters for EfficientNet-B0). The grid search started with the highest possible initial learning rate that was numerically stable for each model; lower initial learning rates were then tested to fine tune the schedule. Zero and nonzero decay target levels were tested for each initial learning rate. In addition to the initial learning rates and decay targets, we experimented with shortening the cosine decay and incorporating restarts. All cross validation experiments were performed on the three splits with $50\%$ of the data for training. 

For training, a cosine learning rate schedule, as in ~\cite{kornblith2019better} with ADAM optimizer~\cite{kingma2014adam} and batchsize 8 was used. For the learning schedules we use the following abbreviations:  Initial Learning Rates (ILR) and decay target (DT).

The tables below list the various initial learning rates explored during cross validation for each model with pretraining. 
\begin{table}[h]
    \begin{center}
    \begin{small}
    \begin{sc}
    \begin{tabular}{lcccc}
        \toprule    
         Model & \multicolumn{4}{c}{ILR}\\
         \midrule
         MobileNetV2-0.35 & 1e-2 & 5e-3 & 1e-3 & 5e-4\\
         MobileNetV2-1.0 & 1e-2 & 5e-3 & 1e-3 & 5e-4\\
         ResNet-50 & 1e-3 & 5e-4 & 1e-4 & 5e-5\\
         EfficientNet-B0 & 2.5e-3 & 1e-3 & 7.5e-4 & 5e-4\\
         EfficientNet-B3 & 1e-3 & 5e-4 & 1e-4 & 5e-5\\
         EfficientNet-B5 & 5e-4 & 1e-4 & &\\
         \bottomrule
    \end{tabular}
    \end{sc}
    \end{small}
    \end{center}
\end{table}

For the ImageNet pretrained case, the learning rate schedule without restarts was optimal on out of domain data, and the resulting optimal parameters are as follows:
\begin{table}[h]
    \begin{center}
    \begin{small}
    \begin{sc}
    \begin{tabular}{lcc}
        \toprule     
         Models & \multicolumn{2}{c}{ILR \& DT}\\
         \midrule
         MobileNetV2s 0.35, 0.5 & 1e-2 & 0 \\
         MobileNetV2s 0.75, 1.0 & 1e-2 & 1e-4\\
         ResNets 50, 101 & 1e-4 & 1e-5\\
         EfficientNets B0, B1 & 5e-4 & 1e-5\\
         EfficientNets B2,B3,B4 & 5e-4 & 0\\
         EfficientNets B5,B6 & 5e-4 & 1e-5\\
         \bottomrule
    \end{tabular}
    \end{sc}
    \end{small}
    \end{center}
    \vspace{-10pt}
\end{table}

The initial learning rates explored for the from scratch models during cross validation are as follows:
\begin{table}[h]
        \begin{center}
    \begin{small}
    \begin{sc}
    \begin{tabular}{lcccc}
        \toprule    
         Model & \multicolumn{4}{c}{ILR}\\
         \midrule
         MobileNetV2 0.35 & 1e-2 & 5e-3 & 1e-3 & 5e-4\\
         MobileNetV2 1.0 & 1e-1 & 1e-2 & 1e-3 & 1e-4\\
         ResNet 50 & 1e-3 & 5e-4 & 1e-4 & 5e-5\\
         EfficientNet-B0 & 1e-3 & 5e-4 & 1e-4 & 5e-5\\
         EfficientNet-B3 & 1e-3 & 5e-4 & 1e-4 & 5e-5\\
         \bottomrule
    \end{tabular}
    \end{sc}
    \end{small}
    \end{center}
    \vspace{-10pt}
\end{table}

For models trained from scratch, we found that using restarts lead to the best performance on out of domain data. The optimal learning rates found during the search are as follows:
\begin{table}[!h]
    \begin{center}
    \begin{small}
    \begin{sc}
    \begin{tabular}{lcc}
        \toprule    
         Models & \multicolumn{2}{c}{ILR \& DT} \\
         \midrule
         MobileNetV2s 0.35, 0.5 & 5e-2 & 5e-3 \\
         MobileNetV2s 0.75, 1.0 & 1e-2 & 0\\
         ResNet 50 & 5e-4 & 5e-5\\
         EfficientNets B0, B3 & 1e-3 & 0\\
         \bottomrule
        \end{tabular}
    \end{sc}
    \end{small}
    \end{center}
    \vspace{-10pt}
\end{table}

\section{Baseline Performance on Horse-30}
For comparison to Horse-10, we provide the train and test normalized errors for models trained on Horse-30. Here, Horse-30 was split into 3 shuffles each containing a train/test split of 50\% of the horse images. Compared to Horse-10, we train these models for twice as long (60,000 iterations) but with the same cross-validated cosine schedules from Horse-10. Errors below are averaged over the three shuffles. \\
\begin{table}[!h]
    \begin{center}
    \begin{small}
    \begin{sc}
    \begin{tabular}{lcccc}
        \toprule
{} & \multicolumn{2}{l}{Horse-10 Errors} & \multicolumn{2}{l}{Horse-30 Errors} \\
Models &    Train &    Test &    Train &    Test \\
\midrule
MobileNetV2 0.35 &   0.1342 &  0.1390 &   0.1545 &  0.1595 \\
ResNet 50        &   0.0742 &  0.0815 &   0.0772 &  0.0825 \\
EfficientNet-B4  &   0.0598 &  0.0686 &   0.0672 &  0.0750 \\
         \bottomrule
        \end{tabular}
    \end{sc}
    \end{small}
    \end{center}
    \vspace{-10pt}
\end{table}

\section{Performance (PCK per bodypart) for all networks on Horse-10}

The tables below show the PCK for several bodyparts for all backbones that we considered. They complete the abridged tables in the main text (Table 2 and 3) 
Thereby the bodyparts are abbreviated as follows: (FF=front foot; HF = Hind foot; HH = Hind Hock).

\begin{table}[ht]
\centering
\caption{PCK@0.3 (\%) for several bodyparts and all evaluated architectures on within domain horses.}
\label{PCK-tableperbptwithinFULL} \small
\begin{tabular}{lcccccccccccc}
\toprule
{} &    Nose &    Eye &  Shoulder &  Wither &  Elbow &  NearFF &  OffFF &    Hip &  NearHH &  NearHF &  OffHF  \\
\midrule
MobileNetV2 0.35 &   90.7 &   94.1 &      97.6 &    96.9 &   96.7 &           92.3 &          93.7 &   96.4 &          94.1 &          94.2 &         92.5  \\
MobileNetV2 0.5  &   94.1 &   96.1 &      99.2 &    98.3 &   98.0 &           93.8 &          95.4 &   96.7 &          97.2 &          97.2 &         97.0  \\
MobileNetV2 0.75 &   96.0 &   97.5 &      99.2 &    98.0 &   99.0 &           96.6 &          96.8 &   98.8 &          97.6 &          98.0 &         97.4  \\
MobileNetV2 1.0  &   97.7 &   98.8 &      99.7 &    99.1 &   99.0 &           97.6 &          97.3 &   99.4 &          98.4 &          98.5 &         98.9  \\
ResNet 50        &   99.9 &  100.0 &      99.8 &    99.9 &   99.8 &           99.8 &          99.6 &   99.9 &          99.9 &          99.6 &         99.8  \\
ResNet 101       &   99.9 &  100.0 &      99.9 &    99.8 &   99.9 &           99.8 &          99.7 &   99.8 &          99.9 &          99.7 &         99.9  \\
EfficientNet-B0  &   99.7 &   99.9 &     100.0 &    99.9 &  100.0 &           99.6 &          99.5 &  100.0 &          99.9 &          99.7 &         99.7  \\
EfficientNet-B1  &   99.8 &   99.9 &     100.0 &    99.8 &   99.9 &           99.5 &          99.8 &  100.0 &          99.8 &          99.8 &         99.8  \\
EfficientNet-B2  &   99.9 &   99.9 &     100.0 &    99.9 &  100.0 &           99.8 &          99.7 &   99.9 &          99.8 &          99.7 &         99.7  \\
EfficientNet-B3  &   99.9 &   99.9 &      99.9 &    99.9 &   99.9 &           99.7 &          99.6 &   99.7 &          99.8 &          99.6 &         99.9  \\
EfficientNet-B4  &  100.0 &  100.0 &      99.9 &    99.8 &   99.9 &           99.6 &          99.7 &   99.9 &          99.7 &          99.8 &         99.8  \\
EfficientNet-B5  &   99.9 &   99.9 &     100.0 &    99.9 &  100.0 &           99.7 &          99.8 &   99.6 &          99.8 &          99.8 &         99.9  \\
EfficientNet-B6  &   99.9 &   99.9 &      99.9 &    99.8 &  100.0 &           99.8 &          99.9 &   99.8 &          99.8 &          99.7 &         99.8 \\
\bottomrule
\end{tabular}
\end{table}

\begin{table*}[ht]
\caption{PCK@0.3 (\%) for several bodyparts and all architectures on out-of-domain horses.}
\label{PCK-tableperbptacrossFULL} \small
\centering
\begin{tabular}{lcccccccccccc}
\toprule
{} &    Nose &    Eye &  Shoulder &  Wither &  Elbow &  NearFF &  OffFF &    Hip &  NearHH &  NearHF &  OffHF  \\
\midrule
MobileNetV2 0.35 &  45.6 &  53.1 &      65.5 &    68.0 &   69.1 &           56.4 &          57.6 &  65.9 &          65.9 &          60.5 &         62.5  \\
MobileNetV2 0.5  &  52.7 &  61.0 &      76.7 &    69.7 &   78.3 &           62.9 &          65.4 &  73.6 &          70.8 &          68.1 &         69.7  \\
MobileNetV2 0.75 &  54.2 &  65.6 &      78.3 &    73.2 &   80.5 &           67.3 &          68.9 &  80.0 &          74.1 &          70.5 &         70.2  \\
MobileNetV2 1.0  &  59.0 &  67.2 &      83.8 &    79.7 &   84.0 &           70.1 &          72.1 &  82.0 &          79.9 &          76.0 &         76.7  \\
ResNet 50        &  68.2 &  73.6 &      85.4 &    85.8 &   88.1 &           72.6 &          70.2 &  89.2 &          85.7 &          77.0 &         74.1  \\
ResNet 101       &  67.7 &  72.4 &      87.6 &    86.0 &   89.0 &           79.9 &          78.0 &  92.6 &          87.2 &          83.4 &         80.0  \\
EfficientNet-B0  &  60.3 &  62.5 &      84.9 &    84.6 &   87.2 &           77.0 &          75.4 &  86.7 &          86.7 &          79.6 &         79.4  \\
EfficientNet-B1  &  67.4 &  71.5 &      85.9 &    85.7 &   89.6 &           80.0 &          81.1 &  86.7 &          88.4 &          81.8 &         81.6  \\
EfficientNet-B2  &  68.7 &  74.8 &      84.5 &    85.2 &   89.2 &           79.7 &          80.9 &  88.1 &          88.0 &          82.3 &         81.7 \\
EfficientNet-B3  &  71.7 &  76.6 &      88.6 &    88.7 &   92.0 &           80.4 &          81.8 &  90.6 &          90.8 &          85.0 &         83.6  \\
EfficientNet-B4  &  71.1 &  75.8 &      88.1 &    87.4 &   91.8 &           83.3 &          82.9 &  90.8 &          90.3 &          86.7 &         85.5  \\
EfficientNet-B5  &  74.8 &  79.5 &      89.6 &    89.5 &   93.5 &           82.2 &          84.1 &  91.8 &          90.9 &          86.6 &         85.2  \\
EfficientNet-B6  &  74.7 &  79.7 &      90.3 &    89.8 &   92.8 &           83.6 &          84.4 &  92.1 &          92.1 &          87.8 &         85.3  \\
\bottomrule
\end{tabular}
\end{table*}

\newpage
\clearpage

\section{CKA analysis of training \&  trained vs. from scratch networks}

\begin{figure*}[h]
\begin{center}
\includegraphics[width=.9\textwidth]{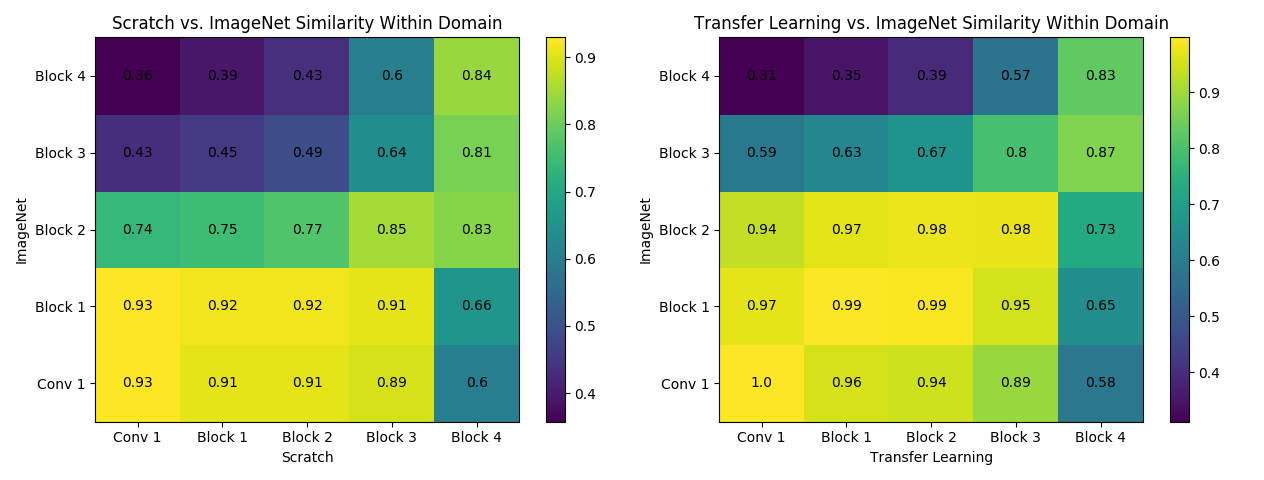}
\end{center}
\caption{CKA comparison of representations for task-training vs. ImageNet trained (no task training) for ResNet-50. Left: Linear CKA on within domain horses (used for training) when trained from scratch vs. plain ImageNet trained (no horse pose estimation task training). Right: Same, but for Transfer Learning vs. from ImageNet. Matrices are the averages over the three splits. In short, task training changes representations.}
\label{fig:TaskvsImageNet}
\end{figure*}

Figure~\ref{fig:TaskvsImageNet} shows a linear centered kernel alignment (CKA)~\cite{kornblith2019similarity} comparison of representations for task-training vs. ImageNet trained (no task training) for ResNet-50

\begin{figure*}
\begin{center}
\includegraphics[width=\textwidth]{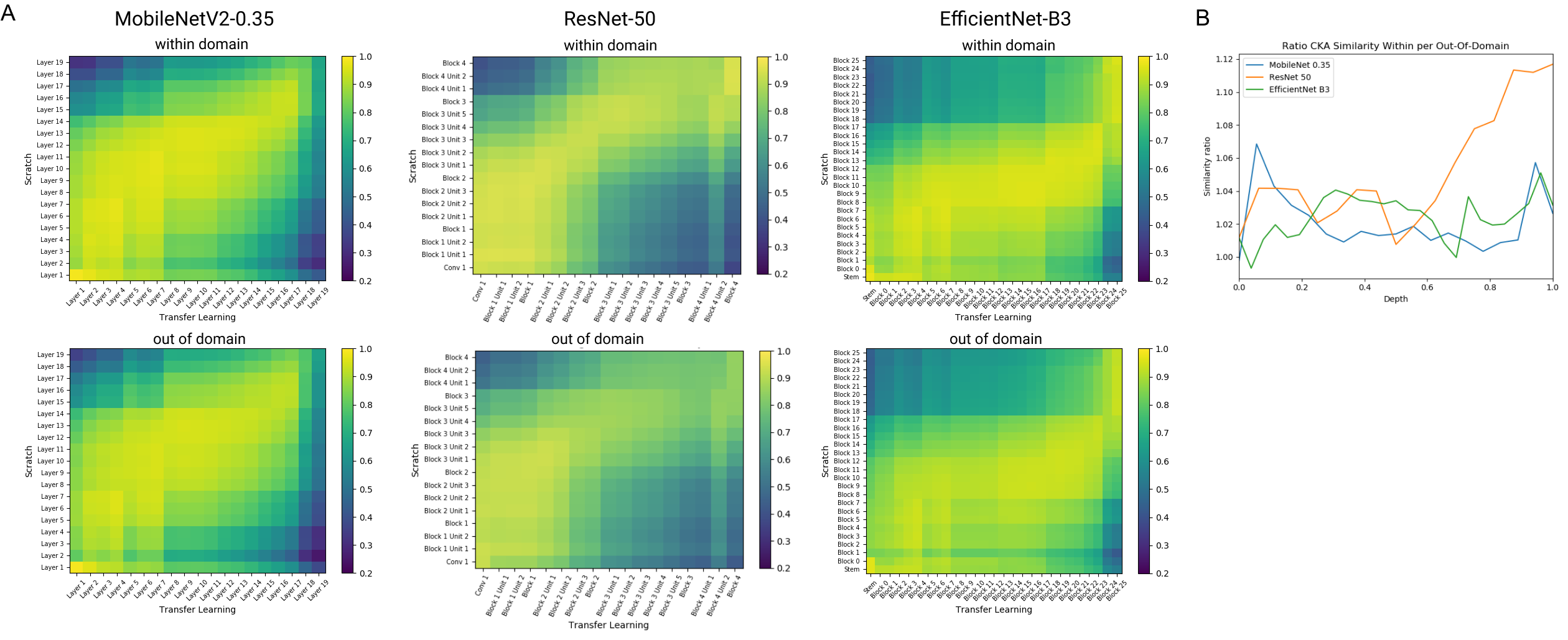}
\end{center}
\caption{\textbf{CKA comparison of representations when trained from scratch vs. from ImageNet initialization.} {\bf A:} Top: Linear CKA between layers of individual networks of different depths on within domain horses (used for training the models). Bottom: Same, but for out-of-domain horses (not used in training). Matrices are the averages over the three splits. {\bf B:} Quantification of similarity ratio plotted against depth of the networks.}
\label{fig:SimPLotsupp}
\vspace{-10pt}
\end{figure*}

\newpage 

\section{Results of within domain performance on Animal Pose}

In Main Figure 4 we show the performance when we train on only 1 species and testing on another (or all without cow/sheep vs. sheep.cow). Here, as a baseline we report the performance within domain, i.e. for each out-of-domain test species (cow and sheep) we trained on 90\% of the cow data and tested on 10\% cow data (see tables~\ref{CowOnCow} and~\ref{SheepOnSheep}). 

\begin{table*}[ht]
\caption{Test performance on cow when trained on 90\% of cow data}
\label{CowOnCow} \small
\centering
\begin{tabular}{lc}
\toprule
{} & Normalized Error  \\
\midrule
MobileNetV2 0.35 & 0.136\\
MobileNetV2 1.0 & 0.093\\
ResNet 50 & 0.062\\
EfficientNet-B0 & 0.060\\
EfficientNet-B3 & 0.054\\
\bottomrule
\end{tabular}
\end{table*}

\begin{table*}[ht]
\caption{Test performance on sheep when trained on 90\% of sheep data}
\label{SheepOnSheep} \small
\centering
\begin{tabular}{lc}
\toprule
{} & Normalized Error  \\
\midrule
MobileNetV2 0.35 & 0.385\\
MobileNetV2 1.0 & 0.248\\
ResNet 50 & 0.186\\
EfficientNet-B0 & 0.124\\
EfficientNet-B3 & 0.159\\
\bottomrule
\end{tabular}
\end{table*}


\section{Full results on Horse-C}

We show the full set of results for the Horse-C benchmark. We compute the corruptions proposed by Hendrycks et al.~\cite{hendrycks2019benchmarking} using the image corruptions library proposed by Michaelis et al.~\cite{michaelis2019dragon}.

The original Horse-30 dataset is processed once for each of the corruptions and severities. In total, Horse-C is comprised of 75 evaluation settings with 8,114 images each, yielding a total of 608,550 images. For a visual impression of the impact of different corruptions and severities, see Figures~\ref{fig:noise}--\ref{fig:digital}.
\input{horse_c_images}

For evaluation, we consider MobileNetV2-0.35, MobileNetV2-1.0, ResNet-50 and the B0 and B3 variants of EfficientNet.
All models are either trained on Horse-10 from scratch or pre-trained on ImageNet and fine-tuned to Horse-10, using the three validation splits used throughout the paper.
In contrast to our other experiments, we now fine-tune the BatchNorm layers for these models. 
For both the w.d. and o.o.d. settings, this yields comparable performance, but enables us to use the batch adaptation technique proposed by Schneider, Rusak et al.~\cite{Schneider2020inc} during evaluation on Horse-C, allowing a better estimate of model robustness.

On the clean data, using batch norm adaptation yields slightly improved performance for MobileNetV2s on clean within-domain data and deteriorates performance for EfficientNet models. Performance on clean ood. data is improved of all model variants when training from scratch, and improved for MobileNets and ResNets when using pre-trained weights.

We evaluate the normalized errors for the non-adapted model (Base) and after estimating corrected batch normalization statistics (Adapt).
The corrected statistics are estimated for each horse identity and corruption as proposed in~\cite{Schneider2020inc}. We average the normalized metrics across shuffles (and horses as usual). We present the full results for a pre-trained ResNet50 model for all four corruption classes in Tables~\ref{tbl:noise-blur} and \ref{tbl:weather-digital} and contrast this to the within-domain/out-of-domain evaluation setting in Table~\ref{tbl:identity}.

For the ResNet50 model considered in detail, we find that batch normalization helps most for noise and weather corruptions, where we typically found improvements of $60-90\%$ and of  $30-70\%$, respectively. In contrast, blur corruptions and digital corruptions (apart from contrast, defocus blur) saw more modest improvements. It is notable that some of the corruptions---such as elastic transform or pixelation---likely also impact the ground truth posture. 

Batch norm adaptation slightly improves the prediction performance when evaluating on different horse identities, but fails to close the gap between the w.d. and ood. setting.
In contrast, batch adaptation considerably improves prediction performance on all considered common corruptions.

In summary, we provide an extensive suite of benchmarks for pose estimation and our experiments suggest that domain shift induced by different individuals is difficult in nature (as it is difficult to fix). This further highlights the importance of benchmarks such as Horse-10.
Full results for other model variants are depicted in Table~\ref{tab:supplement-horsec-overview} and Table~\ref{tab:horsec-full}. We report average scores on Horse-C all models in the main text.


\input{horse_c}

\section{Inference Speed Benchmarking}

\begin{figure}[h]
\begin{center}
\includegraphics[width=.95\textwidth]{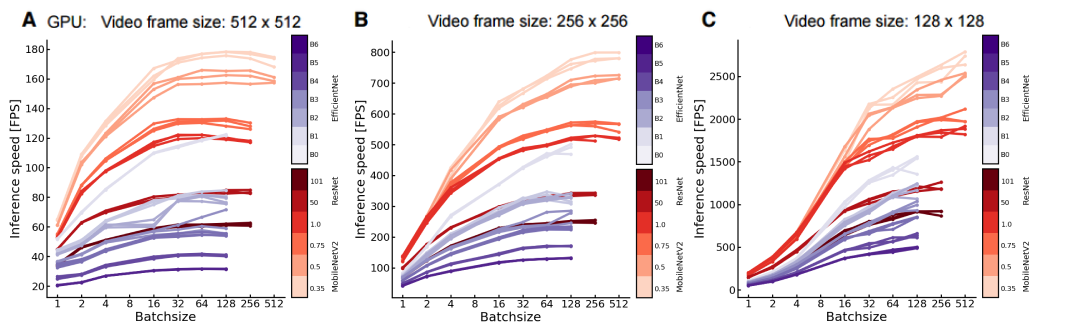}
\end{center}
\caption{{\bf Speed Benchmarking for MobileNetV2s, ResNets and EfficientNets:} Inference speed for videos of different dimensions for all the architectures. {\bf A-C:} FPS vs. batchsize, with video frame sizes as stated in the title. Three splits are shown for each network. MobileNetV2 gives a more than 2X speed improvement (over ResNet-50) for offline processing and about $40\%$ for batchsize=1 on a Titan RTX GPU.}
\label{fig:speed}
\end{figure}

We introduced new DeepLabCut variants that can achieve high accuracy but with higher speed than the original ResNet backbone~\cite{mathis2018deeplabcut}. Here we provide a simple benchmark to document how fast the EfficientNet and MobileNetv2 backbones are (Figure~\ref{fig:speed}). 
We evaluated the inference speed for one video with $11,178$ frames at resolutions $512\times512$, $256\times256$ and $128\times128$. We used batch sizes: $[1, 2, 4, 16, 32, 128, 256, 512]$, and ran all models for all 3 (training set shuffles) trained with $50\%$ of the data in a pseudo random order on a NVIDIA Titan RTX. We also updated the inference code from its numpy implementation \cite{MathisWarren2018speed} to TensorFlow, which brings a $2-10\%$ gain in speed. 

\end{document}

%% file: horse_c_images.tex
\newcommand{\incscale}{.8}

\begin{figure}[h]
\centering
\includegraphics[width=\incscale\textwidth]{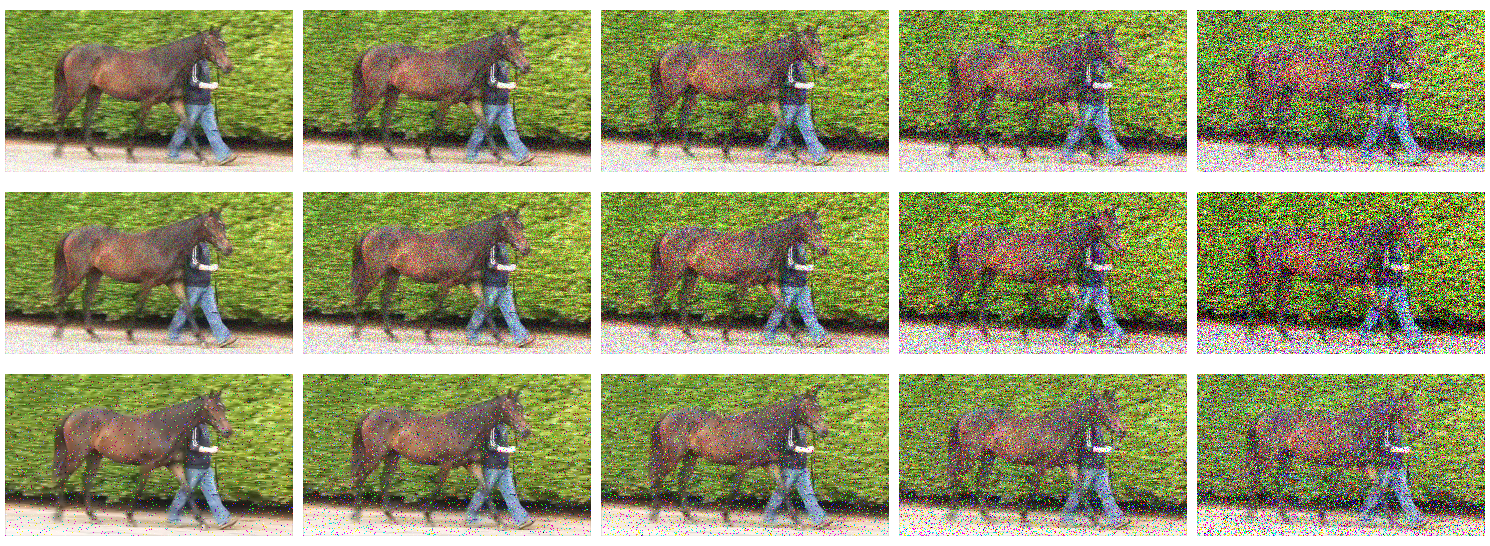}
\caption{Noise corruptions for all five different severities (1 to 5, left to right). Top to bottom: Gaussian Noise, Shot Noise, Impulse Noise.
}
\label{fig:noise}
\end{figure}

\begin{figure}[hp]
\centering
\includegraphics[width=\incscale\textwidth]{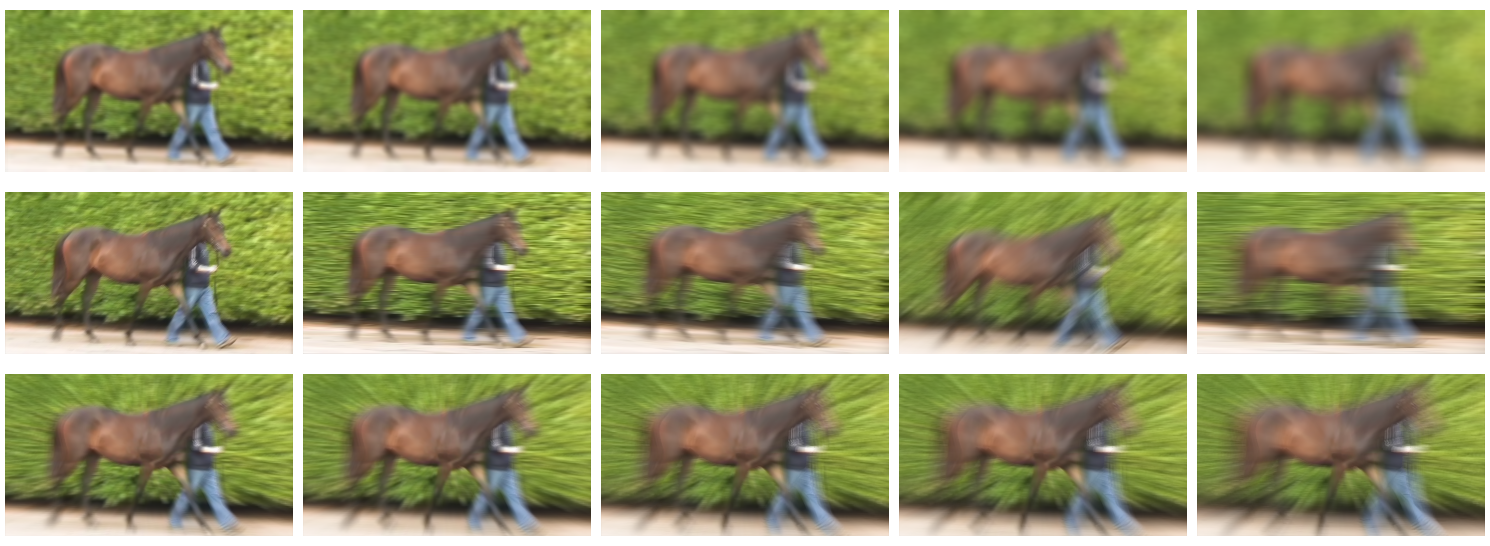}
\caption{Blur corruptions for all five different severities (1 to 5, left to right). Top to bottom: 
Defocus Blur, Motion Blur, Zoom Blur
}
\label{fig:blur}
\end{figure}

\begin{figure}[hp]
\centering
\includegraphics[width=\incscale\textwidth]{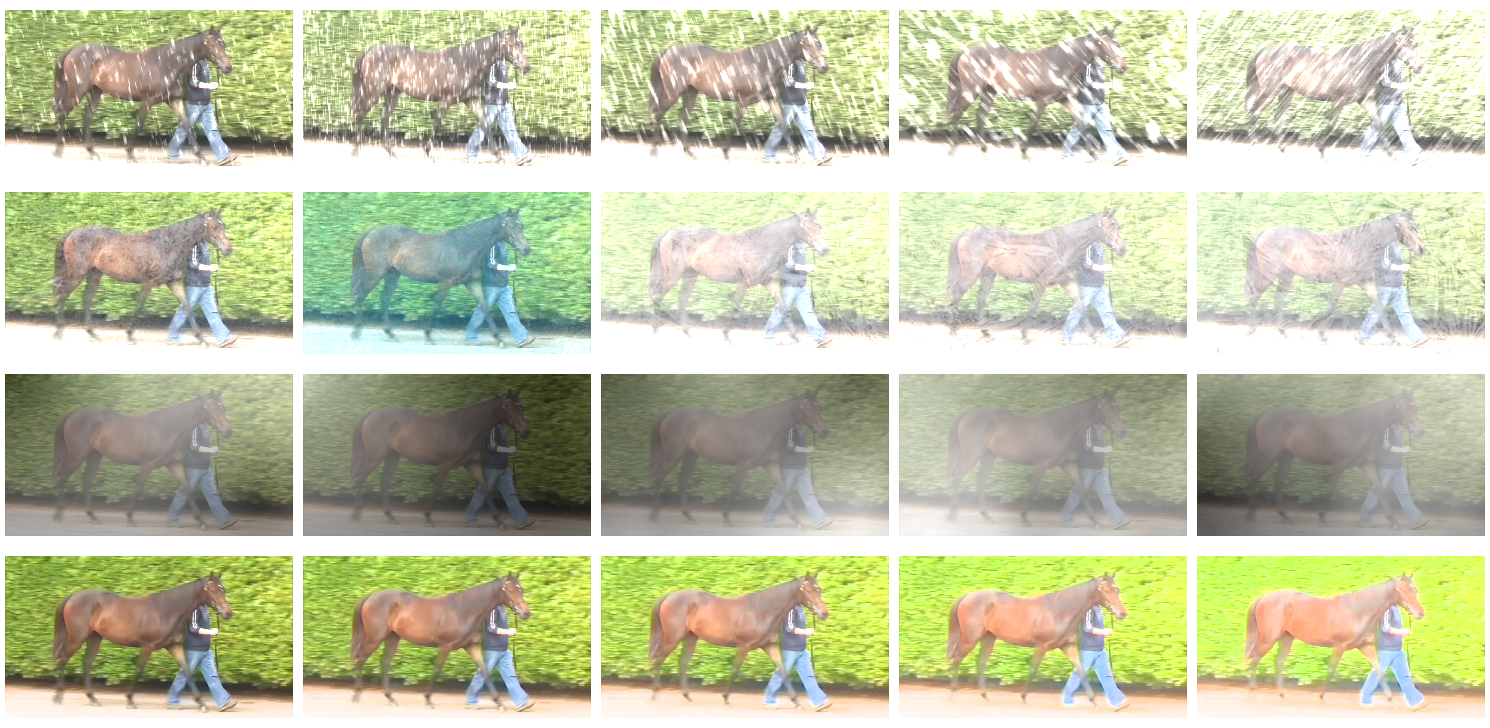}
\caption{Weather corruptions for all five different severities (1 to 5, left to right). Top to bottom: 
Snow, Frost, Fog, Brightness
}
\label{fig:weather}
\end{figure}

\begin{figure}[hp]
\centering
\includegraphics[width=\incscale\textwidth]{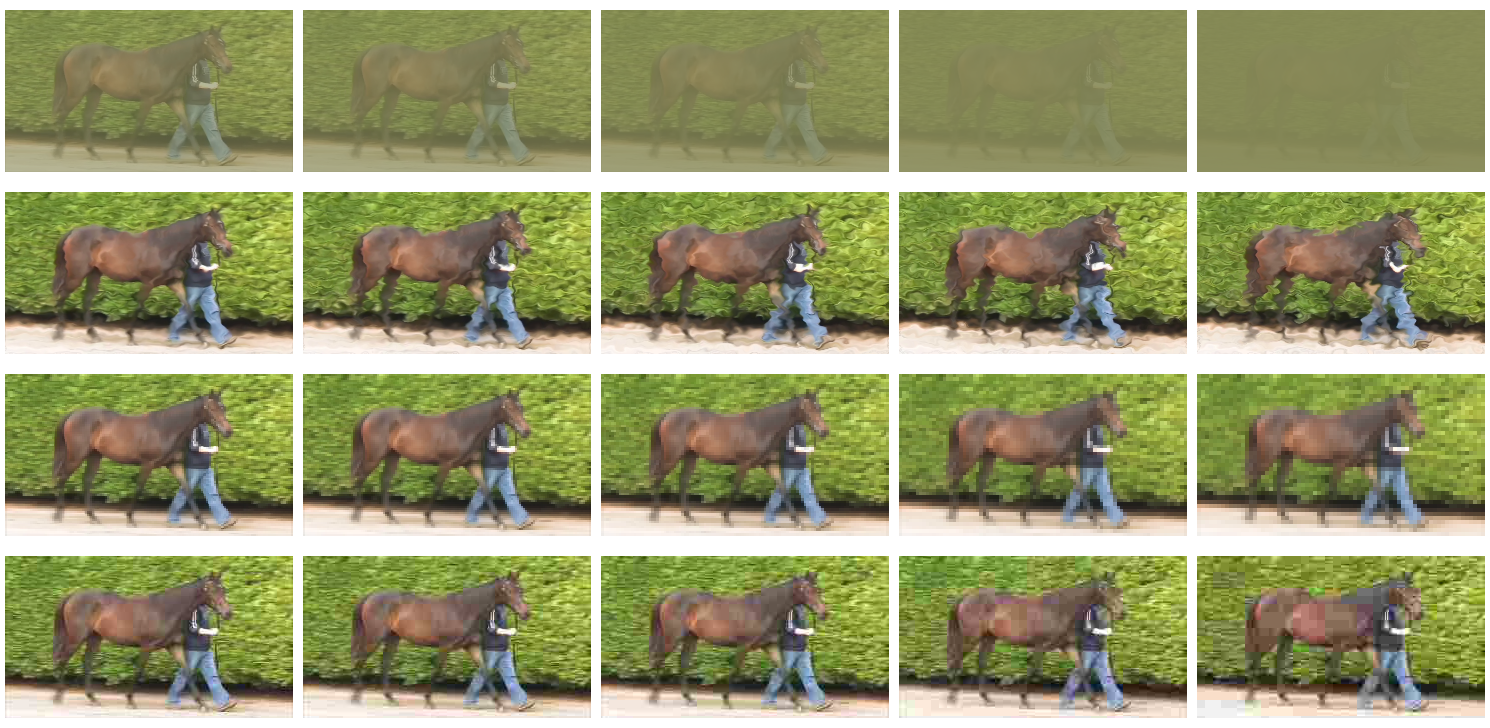}
\caption{Digital corruptions for all five different severities (1 to 5, left to right). Top to bottom: 
Contrast, Elastic Transform, Pixelate, Jpeg Compression
}
\label{fig:digital}
\end{figure}

%% file: horse_c.tex
\begin{table}[]
\caption{Summary results for evaluation of all models on the Horse-C dataset. Results are averaged across all five severities and three validation splits of the data. Adaptive batch normalization (adapt) is crucial for attaining good performance compared to fixing the statistics during evaluation (base). Best viewed in the digital version.}
\centering
\footnotesize
\setlength\tabcolsep{3pt}
\begin{tabular}{lrrrrrrrrrrrrrrrrrrrr}
\toprule
Net Type & \multicolumn{4}{l}{mobilenet\_v2\_0.35} & \multicolumn{4}{l}{mobilenet\_v2\_1.0} & \multicolumn{4}{l}{resnet\_50} & \multicolumn{4}{l}{efficientnet-b0} & \multicolumn{4}{l}{efficientnet-b3} \\
Pretrained & \multicolumn{2}{l}{False} & \multicolumn{2}{l}{True} & \multicolumn{2}{l}{False} & \multicolumn{2}{l}{True} & \multicolumn{2}{l}{False} & \multicolumn{2}{l}{True} & \multicolumn{2}{l}{False} & \multicolumn{2}{l}{True} & \multicolumn{2}{l}{False} & \multicolumn{2}{l}{True} \\
Condition &             adapt &  base & adapt &  base &            adapt &  base & adapt &  base &     adapt &  base & adapt &  base &           adapt &  base & adapt &  base &           adapt &  base & adapt &  base \\
Corruption        &                   &       &       &       &                  &       &       &       &           &       &       &       &                 &       &       &       &                 &       &       &       \\
\midrule
brightness        &              0.34 &  1.76 &  0.29 &  0.87 &             0.27 &  1.67 &  0.21 &  0.94 &      0.33 &  1.29 &  0.17 &  0.24 &            0.40 &  1.40 &  0.19 &  0.72 &            0.33 &  1.38 &  0.19 &  0.70 \\
contrast          &              0.47 &  8.02 &  0.30 &  4.78 &             0.36 &  8.63 &  0.22 &  4.93 &      0.38 &  8.57 &  0.18 &  2.41 &            0.41 &  6.95 &  0.20 &  4.01 &            0.37 &  7.94 &  0.20 &  3.43 \\
defocus\_blur      &              0.81 &  3.22 &  0.69 &  2.20 &             0.61 &  3.51 &  0.66 &  3.35 &      0.83 &  3.44 &  0.60 &  1.54 &            0.67 &  2.05 &  0.51 &  2.54 &            0.71 &  2.64 &  0.54 &  2.01 \\
elastic\_transform &              0.38 &  0.96 &  0.36 &  0.83 &             0.32 &  0.96 &  0.32 &  0.92 &      0.35 &  0.50 &  0.26 &  0.29 &            0.39 &  0.91 &  0.29 &  0.76 &            0.38 &  0.90 &  0.29 &  0.77 \\
fog               &              1.57 &  6.62 &  0.41 &  1.55 &             1.17 &  7.20 &  0.30 &  2.23 &      1.09 &  7.51 &  0.26 &  0.56 &            1.63 &  5.31 &  0.27 &  1.15 &            1.28 &  6.80 &  0.25 &  1.11 \\
frost             &              2.27 &  6.74 &  1.10 &  3.78 &             1.97 &  6.81 &  1.00 &  3.84 &      1.68 &  6.44 &  0.60 &  1.80 &            1.39 &  6.44 &  0.71 &  2.91 &            1.43 &  7.04 &  0.67 &  2.43 \\
gaussian\_noise    &              2.65 &  5.90 &  1.68 &  6.98 &             2.11 &  6.19 &  1.91 &  7.51 &      0.97 &  3.53 &  0.82 &  5.25 &            1.65 &  5.13 &  1.22 &  5.77 &            1.71 &  5.82 &  1.25 &  5.89 \\
glass\_blur        &              0.60 &  2.03 &  0.63 &  1.57 &             0.50 &  2.01 &  0.67 &  2.34 &      0.54 &  1.69 &  0.50 &  0.95 &            0.53 &  1.35 &  0.53 &  1.56 &            0.56 &  1.45 &  0.59 &  1.39 \\
impulse\_noise     &              2.36 &  5.75 &  1.73 &  6.88 &             1.86 &  6.07 &  1.91 &  7.46 &      0.83 &  3.47 &  0.81 &  5.56 &            1.45 &  4.83 &  0.89 &  5.46 &            1.46 &  5.80 &  0.86 &  5.71 \\
jpeg\_compression  &              0.64 &  1.32 &  0.52 &  1.12 &             0.50 &  1.49 &  0.48 &  1.30 &      0.39 &  0.62 &  0.34 &  0.39 &            0.51 &  1.10 &  0.43 &  1.06 &            0.47 &  1.13 &  0.45 &  1.00 \\
motion\_blur       &              0.83 &  2.81 &  0.68 &  1.84 &             0.73 &  2.99 &  0.68 &  2.69 &      0.80 &  2.29 &  0.56 &  1.08 &            0.68 &  1.88 &  0.56 &  1.72 &            0.66 &  2.07 &  0.56 &  1.66 \\
none              &              0.30 &  0.88 &  0.26 &  0.72 &             0.25 &  0.87 &  0.20 &  0.73 &      0.27 &  0.40 &  0.17 &  0.19 &            0.33 &  0.84 &  0.18 &  0.66 &            0.30 &  0.83 &  0.18 &  0.66 \\
pixelate          &              0.34 &  0.99 &  0.33 &  0.84 &             0.28 &  0.96 &  0.28 &  0.99 &      0.31 &  0.47 &  0.23 &  0.28 &            0.35 &  0.89 &  0.27 &  0.78 &            0.33 &  0.86 &  0.27 &  0.76 \\
shot\_noise        &              2.27 &  5.31 &  1.29 &  6.52 &             1.65 &  5.57 &  1.29 &  6.95 &      0.72 &  2.83 &  0.63 &  4.40 &            1.28 &  4.55 &  0.82 &  4.94 &            1.32 &  5.63 &  0.80 &  4.90 \\
snow              &              0.89 &  4.14 &  0.82 &  2.55 &             0.75 &  4.38 &  0.76 &  3.55 &      0.71 &  4.89 &  0.46 &  1.69 &            0.70 &  3.51 &  0.53 &  1.75 &            0.63 &  4.32 &  0.51 &  1.79 \\
zoom\_blur         &              0.98 &  2.34 &  0.82 &  1.74 &             0.88 &  2.58 &  0.89 &  2.39 &      0.93 &  2.16 &  0.69 &  1.11 &            0.93 &  1.75 &  0.70 &  1.60 &            1.02 &  1.95 &  0.71 &  1.56 \\
\bottomrule
\end{tabular}
\label{tab:supplement-horsec-overview}
\end{table}

\begin{table}[]
\caption{Full result table on Horse-C. All results are averaged across the three validation splits. ``none'' denotes the uncorrupted Horse-10 dataset. Best viewed in the digital version.}
\label{tab:horsec-full}
    \scriptsize
    \centering
    \setlength\tabcolsep{3pt}
    \begin{tabular}{llrrrrrrrrrrrrrrrrrrrr}
\toprule
          & Net Type & \multicolumn{4}{l}{mobilenet\_v2\_0.35} & \multicolumn{4}{l}{mobilenet\_v2\_1.0} & \multicolumn{4}{l}{resnet\_50} & \multicolumn{4}{l}{efficientnet-b0} & \multicolumn{4}{l}{efficientnet-b3} \\
          & Pretrained & \multicolumn{2}{l}{False} & \multicolumn{2}{l}{True} & \multicolumn{2}{l}{False} & \multicolumn{2}{l}{True} & \multicolumn{2}{l}{False} & \multicolumn{2}{l}{True} & \multicolumn{2}{l}{False} & \multicolumn{2}{l}{True} & \multicolumn{2}{l}{False} & \multicolumn{2}{l}{True} \\
          & Condition &             adapt &  base & adapt &  base &            adapt &   base & adapt &  base &     adapt &   base & adapt &  base &           adapt &  base & adapt &  base &           adapt &  base & adapt &  base \\
Corruption & Severity &                   &       &       &       &                  &        &       &       &           &        &       &       &                 &       &       &       &                 &       &       &       \\
\midrule
brightness & 1 &              0.30 &  1.02 &  0.26 &  0.75 &             0.25 &   0.96 &  0.20 &  0.78 &      0.29 &   0.44 &  0.16 &  0.20 &            0.34 &  0.92 &  0.18 &  0.67 &            0.30 &  0.86 &  0.17 &  0.68 \\
          & 2 &              0.32 &  1.26 &  0.26 &  0.80 &             0.25 &   1.20 &  0.20 &  0.84 &      0.31 &   0.60 &  0.16 &  0.21 &            0.36 &  1.01 &  0.18 &  0.69 &            0.31 &  0.95 &  0.18 &  0.69 \\
          & 3 &              0.33 &  1.77 &  0.27 &  0.87 &             0.26 &   1.67 &  0.20 &  0.91 &      0.32 &   0.99 &  0.17 &  0.23 &            0.39 &  1.32 &  0.19 &  0.71 &            0.33 &  1.20 &  0.18 &  0.69 \\
          & 4 &              0.35 &  2.20 &  0.30 &  0.93 &             0.27 &   2.14 &  0.22 &  1.02 &      0.34 &   1.71 &  0.18 &  0.26 &            0.42 &  1.67 &  0.20 &  0.74 &            0.34 &  1.61 &  0.20 &  0.70 \\
          & 5 &              0.40 &  2.55 &  0.35 &  1.00 &             0.30 &   2.38 &  0.24 &  1.17 &      0.39 &   2.74 &  0.19 &  0.32 &            0.46 &  2.07 &  0.22 &  0.78 &            0.38 &  2.25 &  0.21 &  0.72 \\
contrast & 1 &              0.31 &  5.23 &  0.26 &  0.96 &             0.25 &   5.25 &  0.20 &  1.02 &      0.27 &   5.64 &  0.17 &  0.27 &            0.33 &  2.50 &  0.18 &  0.80 &            0.30 &  3.87 &  0.18 &  0.73 \\
          & 2 &              0.32 &  6.91 &  0.27 &  1.38 &             0.25 &   7.93 &  0.20 &  1.77 &      0.28 &   7.86 &  0.17 &  0.39 &            0.34 &  5.22 &  0.18 &  1.03 &            0.31 &  7.36 &  0.18 &  0.87 \\
          & 3 &              0.35 &  8.63 &  0.27 &  3.50 &             0.27 &   9.51 &  0.20 &  4.67 &      0.30 &   9.03 &  0.17 &  1.05 &            0.35 &  7.88 &  0.19 &  2.20 &            0.32 &  9.21 &  0.18 &  1.70 \\
          & 4 &              0.48 &  9.54 &  0.29 &  8.13 &             0.36 &  10.22 &  0.22 &  8.17 &      0.38 &  10.05 &  0.18 &  3.90 &            0.40 &  9.48 &  0.20 &  6.85 &            0.36 &  9.60 &  0.20 &  5.66 \\
          & 5 &              0.88 &  9.77 &  0.38 &  9.92 &             0.68 &  10.25 &  0.29 &  9.03 &      0.69 &  10.28 &  0.22 &  6.43 &            0.63 &  9.68 &  0.26 &  9.17 &            0.56 &  9.66 &  0.25 &  8.16 \\
defocus\_blur & 1 &              0.36 &  1.11 &  0.32 &  0.89 &             0.30 &   1.09 &  0.26 &  1.00 &      0.33 &   0.64 &  0.24 &  0.31 &            0.38 &  0.97 &  0.23 &  0.78 &            0.34 &  0.91 &  0.23 &  0.77 \\
          & 2 &              0.41 &  1.48 &  0.39 &  1.14 &             0.35 &   1.39 &  0.31 &  1.50 &      0.39 &   1.11 &  0.29 &  0.40 &            0.42 &  1.11 &  0.28 &  1.02 &            0.39 &  1.05 &  0.28 &  0.91 \\
          & 3 &              0.65 &  3.21 &  0.59 &  1.92 &             0.50 &   3.08 &  0.52 &  3.23 &      0.63 &   3.23 &  0.47 &  0.94 &            0.58 &  1.78 &  0.44 &  2.31 &            0.58 &  2.06 &  0.44 &  1.56 \\
          & 4 &              1.03 &  4.59 &  0.87 &  2.96 &             0.76 &   5.30 &  0.86 &  4.97 &      1.09 &   5.36 &  0.82 &  2.26 &            0.84 &  2.69 &  0.66 &  3.69 &            0.92 &  3.65 &  0.69 &  2.75 \\
          & 5 &              1.60 &  5.69 &  1.28 &  4.11 &             1.12 &   6.69 &  1.36 &  6.06 &      1.72 &   6.83 &  1.19 &  3.77 &            1.13 &  3.68 &  0.96 &  4.88 &            1.30 &  5.52 &  1.05 &  4.05 \\
elastic\_transform & 1 &              0.32 &  0.92 &  0.29 &  0.75 &             0.27 &   0.90 &  0.23 &  0.79 &      0.30 &   0.43 &  0.20 &  0.22 &            0.35 &  0.87 &  0.22 &  0.69 &            0.33 &  0.86 &  0.21 &  0.69 \\
          & 2 &              0.34 &  0.94 &  0.31 &  0.77 &             0.29 &   0.92 &  0.26 &  0.83 &      0.32 &   0.46 &  0.22 &  0.24 &            0.36 &  0.88 &  0.24 &  0.72 &            0.35 &  0.88 &  0.24 &  0.72 \\
          & 3 &              0.38 &  0.96 &  0.35 &  0.82 &             0.31 &   0.95 &  0.31 &  0.91 &      0.35 &   0.50 &  0.25 &  0.28 &            0.39 &  0.91 &  0.29 &  0.75 &            0.37 &  0.90 &  0.28 &  0.76 \\
          & 4 &              0.40 &  0.99 &  0.39 &  0.86 &             0.34 &   0.99 &  0.36 &  0.99 &      0.37 &   0.53 &  0.29 &  0.32 &            0.41 &  0.93 &  0.33 &  0.79 &            0.40 &  0.92 &  0.33 &  0.80 \\
          & 5 &              0.44 &  1.01 &  0.44 &  0.93 &             0.38 &   1.04 &  0.43 &  1.11 &      0.41 &   0.58 &  0.33 &  0.38 &            0.44 &  0.96 &  0.38 &  0.85 &            0.44 &  0.96 &  0.39 &  0.87 \\
fog & 1 &              0.87 &  5.11 &  0.34 &  0.96 &             0.65 &   5.27 &  0.24 &  1.11 &      0.60 &   5.78 &  0.20 &  0.28 &            0.89 &  3.46 &  0.21 &  0.81 &            0.71 &  4.76 &  0.21 &  0.76 \\
          & 2 &              1.26 &  6.42 &  0.36 &  1.21 &             0.94 &   6.93 &  0.26 &  1.57 &      0.82 &   7.33 &  0.22 &  0.33 &            1.30 &  4.89 &  0.23 &  0.96 &            1.03 &  6.53 &  0.22 &  0.86 \\
          & 3 &              1.74 &  7.10 &  0.41 &  1.59 &             1.29 &   7.82 &  0.29 &  2.38 &      1.18 &   8.03 &  0.26 &  0.48 &            1.80 &  5.91 &  0.26 &  1.20 &            1.40 &  7.45 &  0.24 &  1.06 \\
          & 4 &              1.81 &  6.97 &  0.43 &  1.64 &             1.34 &   7.67 &  0.32 &  2.47 &      1.24 &   7.97 &  0.28 &  0.56 &            1.89 &  5.82 &  0.28 &  1.21 &            1.47 &  7.35 &  0.26 &  1.15 \\
          & 5 &              2.17 &  7.49 &  0.51 &  2.33 &             1.65 &   8.29 &  0.40 &  3.60 &      1.60 &   8.46 &  0.36 &  1.13 &            2.27 &  6.45 &  0.35 &  1.58 &            1.77 &  7.93 &  0.32 &  1.74 \\
frost & 1 &              1.02 &  4.11 &  0.46 &  1.32 &             0.73 &   3.81 &  0.37 &  1.30 &      0.65 &   3.24 &  0.26 &  0.37 &            0.58 &  2.94 &  0.34 &  0.88 &            0.56 &  3.52 &  0.33 &  0.82 \\
          & 2 &              1.98 &  6.45 &  0.86 &  2.99 &             1.63 &   6.32 &  0.75 &  2.97 &      1.30 &   6.39 &  0.45 &  1.15 &            1.08 &  6.01 &  0.57 &  1.86 &            1.14 &  7.02 &  0.55 &  1.57 \\
          & 3 &              2.59 &  7.56 &  1.23 &  4.41 &             2.30 &   7.71 &  1.14 &  4.50 &      1.92 &   7.39 &  0.65 &  2.09 &            1.58 &  7.56 &  0.77 &  3.34 &            1.63 &  8.05 &  0.73 &  2.73 \\
          & 4 &              2.75 &  7.66 &  1.31 &  4.70 &             2.41 &   7.92 &  1.21 &  4.81 &      2.06 &   7.48 &  0.71 &  2.30 &            1.67 &  7.60 &  0.84 &  3.69 &            1.77 &  8.16 &  0.79 &  3.02 \\
          & 5 &              3.01 &  7.92 &  1.63 &  5.50 &             2.77 &   8.32 &  1.54 &  5.61 &      2.46 &   7.71 &  0.90 &  3.07 &            2.03 &  8.12 &  1.01 &  4.77 &            2.05 &  8.47 &  0.95 &  3.99 \\
gaussian\_noise & 1 &              0.93 &  3.08 &  0.51 &  3.73 &             0.59 &   2.99 &  0.50 &  4.81 &      0.34 &   0.61 &  0.27 &  0.68 &            0.58 &  1.55 &  0.41 &  1.36 &            0.53 &  1.73 &  0.40 &  1.45 \\
          & 2 &              1.61 &  5.26 &  0.80 &  6.21 &             1.00 &   5.42 &  0.82 &  6.78 &      0.42 &   1.12 &  0.37 &  2.67 &            0.85 &  3.03 &  0.57 &  3.09 &            0.80 &  3.91 &  0.56 &  3.50 \\
          & 3 &              2.63 &  6.60 &  1.41 &  7.96 &             1.87 &   6.91 &  1.56 &  8.31 &      0.63 &   3.13 &  0.57 &  6.12 &            1.45 &  5.25 &  0.93 &  6.59 &            1.43 &  6.37 &  0.91 &  6.66 \\
          & 4 &              3.62 &  7.15 &  2.29 &  8.48 &             2.99 &   7.55 &  2.65 &  8.75 &      1.16 &   5.62 &  1.02 &  8.02 &            2.23 &  7.35 &  1.55 &  8.70 &            2.39 &  7.96 &  1.60 &  8.54 \\
          & 5 &              4.47 &  7.43 &  3.40 &  8.53 &             4.10 &   8.06 &  4.00 &  8.92 &      2.28 &   7.17 &  1.88 &  8.77 &            3.12 &  8.46 &  2.62 &  9.13 &            3.40 &  9.14 &  2.77 &  9.29 \\
glass\_blur & 1 &              0.33 &  0.98 &  0.30 &  0.82 &             0.27 &   0.94 &  0.25 &  0.85 &      0.31 &   0.50 &  0.22 &  0.27 &            0.35 &  0.89 &  0.21 &  0.71 &            0.33 &  0.85 &  0.21 &  0.71 \\
          & 2 &              0.38 &  1.11 &  0.37 &  0.95 &             0.32 &   1.11 &  0.33 &  1.07 &      0.35 &   0.68 &  0.28 &  0.35 &            0.39 &  0.98 &  0.28 &  0.82 &            0.36 &  0.93 &  0.30 &  0.81 \\
          & 3 &              0.53 &  1.54 &  0.58 &  1.38 &             0.47 &   1.45 &  0.62 &  1.83 &      0.51 &   1.19 &  0.48 &  0.67 &            0.52 &  1.14 &  0.49 &  1.15 &            0.52 &  1.13 &  0.55 &  1.13 \\
          & 4 &              0.65 &  2.38 &  0.72 &  1.79 &             0.57 &   2.06 &  0.83 &  2.93 &      0.59 &   1.84 &  0.60 &  1.06 &            0.58 &  1.42 &  0.66 &  1.61 &            0.63 &  1.48 &  0.74 &  1.51 \\
          & 5 &              1.09 &  4.16 &  1.18 &  2.91 &             0.88 &   4.51 &  1.34 &  5.04 &      0.95 &   4.26 &  0.93 &  2.41 &            0.83 &  2.33 &  1.01 &  3.50 &            0.98 &  2.89 &  1.17 &  2.80 \\
impulse\_noise & 1 &              0.92 &  2.86 &  0.56 &  3.69 &             0.58 &   2.80 &  0.56 &  4.72 &      0.38 &   0.76 &  0.36 &  1.43 &            0.63 &  1.66 &  0.33 &  1.12 &            0.56 &  2.08 &  0.31 &  1.05 \\
          & 2 &              1.53 &  5.19 &  0.85 &  6.26 &             1.00 &   5.37 &  0.94 &  6.95 &      0.47 &   1.50 &  0.47 &  3.89 &            0.89 &  3.17 &  0.47 &  2.92 &            0.82 &  4.55 &  0.43 &  3.50 \\
          & 3 &              2.10 &  6.26 &  1.21 &  7.45 &             1.44 &   6.53 &  1.42 &  7.95 &      0.57 &   2.69 &  0.58 &  5.74 &            1.18 &  4.45 &  0.62 &  5.56 &            1.12 &  5.87 &  0.57 &  5.83 \\
          & 4 &              3.18 &  7.06 &  2.36 &  8.42 &             2.61 &   7.55 &  2.67 &  8.75 &      0.99 &   5.38 &  0.97 &  7.98 &            1.90 &  6.82 &  1.12 &  8.60 &            1.93 &  7.60 &  1.06 &  8.76 \\
          & 5 &              4.07 &  7.38 &  3.64 &  8.58 &             3.68 &   8.07 &  3.95 &  8.93 &      1.76 &   6.99 &  1.68 &  8.78 &            2.67 &  8.07 &  1.90 &  9.10 &            2.84 &  8.90 &  1.94 &  9.38 \\
jpeg\_compression & 1 &              0.45 &  1.05 &  0.35 &  0.88 &             0.37 &   1.09 &  0.33 &  0.96 &      0.31 &   0.45 &  0.24 &  0.28 &            0.40 &  0.92 &  0.29 &  0.77 &            0.37 &  0.93 &  0.29 &  0.77 \\
          & 2 &              0.53 &  1.12 &  0.41 &  0.95 &             0.42 &   1.18 &  0.37 &  1.05 &      0.35 &   0.48 &  0.27 &  0.31 &            0.45 &  0.95 &  0.34 &  0.83 &            0.41 &  0.98 &  0.35 &  0.83 \\
          & 3 &              0.66 &  1.16 &  0.51 &  1.03 &             0.49 &   1.25 &  0.43 &  1.14 &      0.37 &   0.50 &  0.30 &  0.33 &            0.48 &  1.01 &  0.39 &  0.91 &            0.44 &  1.07 &  0.39 &  0.91 \\
          & 4 &              0.75 &  1.45 &  0.61 &  1.25 &             0.58 &   1.66 &  0.56 &  1.49 &      0.42 &   0.70 &  0.38 &  0.44 &            0.57 &  1.23 &  0.49 &  1.17 &            0.53 &  1.27 &  0.53 &  1.11 \\
          & 5 &              0.80 &  1.81 &  0.71 &  1.51 &             0.62 &   2.26 &  0.70 &  1.86 &      0.51 &   0.95 &  0.52 &  0.57 &            0.63 &  1.37 &  0.65 &  1.62 &            0.58 &  1.39 &  0.68 &  1.38 \\
motion\_blur & 1 &              0.39 &  1.15 &  0.34 &  0.86 &             0.34 &   1.11 &  0.30 &  0.98 &      0.37 &   0.60 &  0.26 &  0.35 &            0.41 &  1.02 &  0.28 &  0.80 &            0.38 &  0.97 &  0.27 &  0.81 \\
          & 2 &              0.51 &  1.54 &  0.44 &  1.02 &             0.45 &   1.49 &  0.41 &  1.31 &      0.48 &   0.95 &  0.36 &  0.48 &            0.49 &  1.20 &  0.38 &  0.98 &            0.46 &  1.16 &  0.38 &  0.96 \\
          & 3 &              0.73 &  2.69 &  0.61 &  1.47 &             0.63 &   2.59 &  0.60 &  2.27 &      0.70 &   1.90 &  0.51 &  0.77 &            0.64 &  1.67 &  0.53 &  1.41 &            0.61 &  1.69 &  0.53 &  1.36 \\
          & 4 &              1.10 &  3.94 &  0.88 &  2.46 &             0.96 &   4.39 &  0.91 &  3.90 &      1.05 &   3.47 &  0.74 &  1.46 &            0.85 &  2.46 &  0.73 &  2.29 &            0.84 &  2.77 &  0.73 &  2.17 \\
          & 5 &              1.44 &  4.73 &  1.11 &  3.41 &             1.26 &   5.38 &  1.17 &  4.97 &      1.38 &   4.52 &  0.95 &  2.32 &            1.03 &  3.06 &  0.88 &  3.14 &            1.03 &  3.75 &  0.88 &  3.01 \\
pixelate & 1 &              0.30 &  0.92 &  0.27 &  0.74 &             0.25 &   0.89 &  0.21 &  0.78 &      0.28 &   0.41 &  0.18 &  0.21 &            0.33 &  0.85 &  0.20 &  0.69 &            0.31 &  0.84 &  0.19 &  0.69 \\
          & 2 &              0.31 &  0.93 &  0.28 &  0.75 &             0.25 &   0.89 &  0.23 &  0.80 &      0.28 &   0.42 &  0.18 &  0.22 &            0.33 &  0.86 &  0.23 &  0.70 &            0.31 &  0.84 &  0.21 &  0.69 \\
          & 3 &              0.33 &  0.98 &  0.31 &  0.82 &             0.26 &   0.93 &  0.26 &  0.91 &      0.29 &   0.46 &  0.22 &  0.27 &            0.35 &  0.88 &  0.24 &  0.73 &            0.32 &  0.86 &  0.25 &  0.73 \\
          & 4 &              0.37 &  1.04 &  0.36 &  0.91 &             0.30 &   1.02 &  0.34 &  1.13 &      0.32 &   0.50 &  0.26 &  0.34 &            0.37 &  0.91 &  0.31 &  0.83 &            0.35 &  0.88 &  0.31 &  0.80 \\
          & 5 &              0.38 &  1.09 &  0.41 &  0.98 &             0.32 &   1.08 &  0.38 &  1.33 &      0.35 &   0.56 &  0.29 &  0.38 &            0.39 &  0.94 &  0.38 &  0.96 &            0.36 &  0.90 &  0.37 &  0.89 \\
shot\_noise & 1 &              0.69 &  1.98 &  0.44 &  2.43 &             0.43 &   1.86 &  0.40 &  3.21 &      0.31 &   0.52 &  0.23 &  0.41 &            0.46 &  1.14 &  0.32 &  0.97 &            0.43 &  1.24 &  0.30 &  0.98 \\
          & 2 &              1.19 &  4.19 &  0.65 &  5.61 &             0.68 &   4.14 &  0.62 &  5.99 &      0.37 &   0.88 &  0.31 &  1.52 &            0.66 &  2.10 &  0.44 &  1.90 &            0.60 &  2.97 &  0.41 &  2.15 \\
          & 3 &              1.97 &  5.93 &  1.02 &  7.70 &             1.19 &   6.31 &  1.01 &  7.93 &      0.50 &   1.89 &  0.46 &  4.28 &            1.00 &  4.06 &  0.63 &  4.49 &            0.93 &  5.95 &  0.59 &  4.58 \\
          & 4 &              3.38 &  7.07 &  1.87 &  8.41 &             2.49 &   7.55 &  1.84 &  8.77 &      0.92 &   4.68 &  0.86 &  7.45 &            1.82 &  7.17 &  1.10 &  8.33 &            1.89 &  8.46 &  1.09 &  7.82 \\
          & 5 &              4.12 &  7.41 &  2.50 &  8.44 &             3.47 &   7.98 &  2.56 &  8.84 &      1.51 &   6.16 &  1.29 &  8.34 &            2.46 &  8.30 &  1.60 &  9.01 &            2.74 &  9.55 &  1.63 &  8.98 \\
snow & 1 &              0.49 &  1.51 &  0.47 &  1.09 &             0.41 &   1.67 &  0.41 &  1.30 &      0.41 &   1.21 &  0.30 &  0.44 &            0.45 &  1.20 &  0.36 &  0.87 &            0.42 &  1.30 &  0.33 &  0.81 \\
          & 2 &              0.74 &  3.60 &  0.71 &  2.38 &             0.63 &   4.19 &  0.70 &  3.51 &      0.61 &   5.65 &  0.43 &  1.38 &            0.60 &  2.60 &  0.51 &  1.42 &            0.55 &  3.73 &  0.48 &  1.34 \\
          & 3 &              0.85 &  3.89 &  0.77 &  2.16 &             0.72 &   4.32 &  0.70 &  3.12 &      0.66 &   4.64 &  0.45 &  1.27 &            0.69 &  2.78 &  0.51 &  1.38 &            0.61 &  3.71 &  0.50 &  1.44 \\
          & 4 &              1.20 &  5.37 &  1.06 &  3.23 &             1.02 &   5.56 &  0.98 &  4.61 &      0.94 &   6.10 &  0.59 &  2.53 &            0.91 &  4.58 &  0.63 &  2.13 &            0.81 &  5.61 &  0.62 &  2.42 \\
          & 5 &              1.15 &  6.35 &  1.07 &  3.92 &             0.97 &   6.18 &  1.03 &  5.23 &      0.91 &   6.85 &  0.54 &  2.82 &            0.83 &  6.39 &  0.63 &  2.93 &            0.75 &  7.23 &  0.63 &  2.93 \\
zoom\_blur & 1 &              0.67 &  1.69 &  0.58 &  1.16 &             0.59 &   1.65 &  0.58 &  1.50 &      0.63 &   1.20 &  0.45 &  0.56 &            0.64 &  1.28 &  0.45 &  1.04 &            0.67 &  1.29 &  0.46 &  1.05 \\
          & 2 &              0.85 &  2.05 &  0.78 &  1.51 &             0.76 &   2.13 &  0.85 &  2.05 &      0.80 &   1.70 &  0.64 &  0.87 &            0.81 &  1.51 &  0.63 &  1.37 &            0.88 &  1.58 &  0.65 &  1.35 \\
          & 3 &              0.95 &  2.39 &  0.78 &  1.64 &             0.84 &   2.60 &  0.83 &  2.32 &      0.92 &   2.13 &  0.66 &  1.04 &            0.90 &  1.72 &  0.67 &  1.50 &            0.99 &  1.91 &  0.67 &  1.45 \\
          & 4 &              1.11 &  2.65 &  0.93 &  1.98 &             1.00 &   3.02 &  1.05 &  2.87 &      1.08 &   2.62 &  0.85 &  1.40 &            1.06 &  1.94 &  0.84 &  1.86 &            1.19 &  2.27 &  0.85 &  1.83 \\
          & 5 &              1.34 &  2.95 &  1.01 &  2.41 &             1.21 &   3.48 &  1.15 &  3.23 &      1.24 &   3.12 &  0.87 &  1.68 &            1.21 &  2.27 &  0.93 &  2.24 &            1.36 &  2.70 &  0.92 &  2.14 \\
none (Horse-10) & 0 &              0.30 &  0.88 &  0.26 &  0.72 &             0.25 &   0.87 &  0.20 &  0.73 &      0.27 &   0.40 &  0.17 &  0.19 &            0.33 &  0.84 &  0.18 &  0.66 &            0.30 &  0.83 &  0.18 &  0.66 \\
\bottomrule
\end{tabular}
\end{table}

\begin{table*}[hp]
\centering
\small
\caption{Improvements using batch adaptation on the Horse-C Noise and Blur corruption subsets for a pre-trained ResNet50 model.}
\begin{tabular}{llrrrl}
\toprule
Noise           &   &   Base &  Adapt &  $\Delta_\text{abs}$ & $\Delta_\text{rel}$ \\
Corruption & Severity &        &        &                      &                     \\
\midrule
Gaussian Noise & 1 &  0.427 &  0.138 &                0.289 &              67.7\% \\
           & 2 &  2.187 &  0.201 &                1.986 &              90.8\% \\
           & 3 &  5.556 &  0.314 &                5.242 &              94.3\% \\
           & 4 &  7.843 &  0.649 &                7.194 &              91.7\% \\
           & 5 &  8.894 &  1.410 &                7.484 &              84.1\% \\
Impulse Noise & 1 &  1.079 &  0.201 &                0.878 &              81.4\% \\
           & 2 &  3.432 &  0.276 &                3.156 &              92.0\% \\
           & 3 &  5.393 &  0.360 &                5.033 &              93.3\% \\
           & 4 &  7.839 &  0.663 &                7.176 &              91.5\% \\
           & 5 &  8.923 &  1.339 &                7.584 &              85.0\% \\
Shot Noise & 1 &  0.191 &  0.114 &                0.077 &              40.3\% \\
           & 2 &  0.986 &  0.152 &                0.834 &              84.6\% \\
           & 3 &  3.618 &  0.244 &                3.374 &              93.3\% \\
           & 4 &  7.225 &  0.516 &                6.709 &              92.9\% \\
           & 5 &  8.365 &  0.894 &                7.471 &              89.3\% \\
\bottomrule
\end{tabular}
~
\begin{tabular}{llrrrl}
\toprule
Blur          &   &   Base &  Adapt &  $\Delta_\text{abs}$ & $\Delta_\text{rel}$ \\
Corruption & Severity &        &        &                      &                     \\
\midrule
Defocus Blur & 1 &  0.137 &  0.100 &                0.037 &              27.0\% \\
          & 2 &  0.169 &  0.127 &                0.042 &              24.9\% \\
          & 3 &  0.369 &  0.233 &                0.136 &              36.9\% \\
          & 4 &  1.569 &  0.446 &                1.123 &              71.6\% \\
          & 5 &  3.480 &  0.763 &                2.717 &              78.1\% \\
Motion Blur & 1 &  0.213 &  0.160 &                0.053 &              24.9\% \\
          & 2 &  0.290 &  0.224 &                0.066 &              22.8\% \\
          & 3 &  0.340 &  0.335 &                0.005 &               1.5\% \\
          & 4 &  0.864 &  0.501 &                0.363 &              42.0\% \\
          & 5 &  1.596 &  0.645 &                0.951 &              59.6\% \\
Zoom Blur & 1 &  0.331 &  0.288 &                0.043 &              13.0\% \\
          & 2 &  0.536 &  0.436 &                0.100 &              18.7\% \\
          & 3 &  0.654 &  0.467 &                0.187 &              28.6\% \\
          & 4 &  0.974 &  0.620 &                0.354 &              36.3\% \\
          & 5 &  1.217 &  0.640 &                0.577 &              47.4\% \\
\bottomrule
\end{tabular}
\label{tbl:noise-blur}
\end{table*}

\begin{table*}[hp]
\centering
\small
\caption{Improvements using batch adaptation on the Horse-C Weather and Digital corruptions subsets  for a pre-trained ResNet50 model.}
\begin{tabular}{llrrrl}
\toprule
Weather     &   &   Base &  Adapt &  $\Delta_\text{abs}$ & $\Delta_\text{rel}$ \\
Corruption & Severity &        &        &                      &                     \\
\midrule
Brightness & 1 &  0.120 &  0.084 &                0.036 &              30.0\% \\
     & 2 &  0.127 &  0.083 &                0.044 &              34.6\% \\
     & 3 &  0.141 &  0.084 &                0.057 &              40.4\% \\
     & 4 &  0.165 &  0.089 &                0.076 &              46.1\% \\
     & 5 &  0.205 &  0.097 &                0.108 &              52.7\% \\
Fog & 1 &  0.156 &  0.097 &                0.059 &              37.8\% \\
     & 2 &  0.191 &  0.107 &                0.084 &              44.0\% \\
     & 3 &  0.289 &  0.126 &                0.163 &              56.4\% \\
     & 4 &  0.330 &  0.137 &                0.193 &              58.5\% \\
     & 5 &  0.764 &  0.176 &                0.588 &              77.0\% \\
Frost & 1 &  0.193 &  0.125 &                0.068 &              35.2\% \\
     & 2 &  0.672 &  0.249 &                0.423 &              62.9\% \\
     & 3 &  1.447 &  0.393 &                1.054 &              72.8\% \\
     & 4 &  1.680 &  0.449 &                1.231 &              73.3\% \\
     & 5 &  2.375 &  0.573 &                1.802 &              75.9\% \\
Snow & 1 &  0.229 &  0.155 &                0.074 &              32.3\% \\
     & 2 &  0.737 &  0.252 &                0.485 &              65.8\% \\
     & 3 &  0.720 &  0.270 &                0.450 &              62.5\% \\
     & 4 &  1.873 &  0.386 &                1.487 &              79.4\% \\
     & 5 &  2.146 &  0.348 &                1.798 &              83.8\% \\
\bottomrule
\end{tabular}
~
\begin{tabular}{llrrrl}
\toprule
Digital         &   &   Base &  Adapt &  $\Delta_\text{abs}$ & $\Delta_\text{rel}$ \\
Corruption & Severity &        &        &                      &                     \\
\midrule
Contrast & 1 &  0.151 &  0.085 &                0.066 &              43.7\% \\
         & 2 &  0.211 &  0.084 &                0.127 &              60.2\% \\
         & 3 &  0.840 &  0.083 &                0.757 &              90.1\% \\
         & 4 &  3.700 &  0.085 &                3.615 &              97.7\% \\
         & 5 &  6.406 &  0.103 &                6.303 &              98.4\% \\
Elastic Transform & 1 &  0.121 &  0.092 &                0.029 &              24.0\% \\
         & 2 &  0.127 &  0.101 &                0.026 &              20.5\% \\
         & 3 &  0.139 &  0.116 &                0.023 &              16.5\% \\
         & 4 &  0.154 &  0.133 &                0.021 &              13.6\% \\
         & 5 &  0.175 &  0.157 &                0.018 &              10.3\% \\
Jpeg Compression & 1 &  0.136 &  0.108 &                0.028 &              20.6\% \\
         & 2 &  0.152 &  0.128 &                0.024 &              15.8\% \\
         & 3 &  0.170 &  0.138 &                0.032 &              18.8\% \\
         & 4 &  0.216 &  0.189 &                0.027 &              12.5\% \\
         & 5 &  0.305 &  0.276 &                0.029 &               9.5\% \\
Pixelate & 1 &  0.117 &  0.087 &                0.030 &              25.6\% \\
         & 2 &  0.117 &  0.089 &                0.028 &              23.9\% \\
         & 3 &  0.125 &  0.100 &                0.025 &              20.0\% \\
         & 4 &  0.142 &  0.112 &                0.030 &              21.1\% \\
         & 5 &  0.156 &  0.132 &                0.024 &              15.4\% \\
\bottomrule
\end{tabular}
\label{tbl:weather-digital}
\end{table*}

\begin{table*}
\centering
\small
\caption{Small improvements by using batch adaptation on the identity shift task for a pre-trained ResNet50 model. Note that the o.o.d. performance is still substantially worse (higher normalized error) than the within-domain performance.}
\begin{tabular}{lrrrl}
\toprule
     &   Base &  Adapt &  $\Delta_\text{abs}$ & $\Delta_\text{rel}$ \\
\midrule
Identity (wd) &  0.115 &  0.086 &                0.029 &              25.2\%\\
Identity (ood) &  0.271 &  0.247 &                0.024 &               8.9\% \\
\bottomrule
\end{tabular}
\label{tbl:identity}
\end{table*}

\clearpage